# PRET is a few-shot system for pan-cancer recognition without example training

**Yi Li[1,16], Ziyu Ning[2,16], Tianqi Xiang[1], Qixiang Zhang[1], Zhihao Lin[2], Min Yi[3], Feiyan Feng[2], Baozhen Zeng[2], Xuexia Qian[2], Lu Sun[2], Jiace Qin[2], Ling Xiang[2], Chao Fan[2], Tian Qin[2], Qian Wang[2], Xiu-Wu Bian[4], Kun-Hsing Yu[5,6,7,8], Kang Zhang[9,10,11,12,*], Qingling Zhang[2,*], Xiaomeng Li[1,13,14,15,*]**

[1]Department of Electronic and Computer Engineering, The Hong Kong University of Science and Technology, Hong Kong SAR, China.

[2]Department of Pathology, Guangdong Provincial People's Hospital (Guangdong Academy of Medical Sciences), Southern Medical University, Guangzhou, China.

[3]Department of Pathology, the Fifth People's Hospital of Qinghai Province, Qinghai Province Cancer Hospital, Xining, China.

[4]Institute of Pathology and Southwest Cancer Center, the First Affiliated Hospital (Southwest Hospital) and School of Basic Medical Sciences, Army Medical University (Third Military Medical University) and the Key Laboratory of Tumor Immunopathology, The Ministry of Education (Third Military Medical University), Chongqing, China.

[5]Department of Biomedical Informatics, Harvard Medical School, Boston, MA, USA.

[6]Department of Pathology, Brigham and Women's Hospital, Boston, MA, USA.

[7]Harvard Data Science Initiative, Harvard University, Cambridge, MA, USA.

[8]Kempner Institute for the Study of Natural and Artificial Intelligence, Harvard University, Cambridge, MA, USA.

[9]Institute for AI in Medicine, Faculty of Medicine, Macau University of Science and Technology, Macau SAR, China.

[10]Department of General Surgery, Zhejiang Key Laboratory of Intelligent Cancer Biomarker Discovery and Translation, The First Affiliated Hospital; Clinical Data Science Institute, Wenzhou Medical University, Wenzhou, China.

[11]Department of Hepatobiliary Surgery, Zhejiang Key Laboratory of Intelligent Cancer Biomarker Discovery and Translation, The First Affiliated Hospital; Clinical Data Science Institute, Wenzhou Medical University, Wenzhou, China.

[12]Guangzhou National Laboratory, Guangzhou, China.

[13]Department of Computer Science and Engineering, The Hong Kong University of Science and Technology, Hong Kong SAR, China.

[14]Shenzhen-Hong Kong Collaborative Innovation Research Institute, The Hong Kong University of Science and Technology, Shenzhen, China.

[15]Shenzhen Loop Area Institute, Shenzhen, China.

[16]These authors contributed equally: Yi Li, Ziyu Ning. *Corresponding authors: Kang Zhang (kang.zhang@gmail.com); Qingling Zhang (zhangqingling@gdph.org.cn); Xiaomeng Li (eexmli@ust.hk)

## ABSTRACT

Pathological examination stands as the cornerstone in cancer diagnosis, impacting millions worldwide annually. With the shortage of pathologists globally, artificial intelligence (AI) has emerged rapidly to automate the diagnostics process. However, conventional AI models require substantial labeled data for each disease, posing huge challenges in scalability and practicality. Therefore, we introduce PRET (pan-cancer recognition without examples training), a few-shot system to achieve flexible, scalable, and effective cancer recognition across diverse organs, hospitals and tasks without training. Evaluated on 23 international benchmarks comprising 4,484 whole-slide images, our method outperforms existing approaches across 20 tasks, achieving over 97% area under the curve on 15 benchmarks with a maximum improvement of 36.76%. Notably, PRET delivers clinical-grade diagnostic performance in lymph node metastasis detection using only eight slide examples, outperforming 11 pathologists. By offering a flexible and cost-effective solution for pan-cancer recognition, PRET paves the way for accessible and equitable AI-based pathology systems, particularly benefiting minority populations and underserved regions.

## Introduction

Pathological examination, the gold standard for cancer diagnosis, is indispensable for approximately 19.3 million new cancer diagnoses annually worldwide[1]. However, the global scarcity of pathologists[2–4] poses notable challenges, particularly in resource-limited settings. Computational pathology[5,6] based on artificial intelligence (AI) has emerged as a solution for automating or assisting the cancer diagnosis, yet conventional AI algorithms are limited in disease generalizability and rely on extensive annotated data[7,8]. For instance, Campanella et al.[7] trained AI classification models for prostate cancer, skin cancer and lymph node metastases with nearly 45,000 whole-slide images (WSIs). Given that the OncoTree[9] cataloged nearly 900 tumor types, the practical limitation of developing specific models for each disease type becomes evident, particularly in low-resource settings where access to expertise and labeled data is limited.

To overcome these barriers, developing a pan-cancer model capable of recognizing cancers with minimal data is crucial. Pan-cancer recognition aims to diagnose multiple cancers using a single general-purpose model, which differs from task-specific models [7,10,11] or pan-cancer analysis of genetic mutations[12,13] and biomarkers[14]. The advent of pathological foundation models[11,12,15,16] has made pan-cancer representations more accessible. These models derive insights either from self-supervised models [12,16,17] trained on vast amounts of unlabeled images or from visual-language models[11,15] supervised by millions of textual labels. Despite their potential, these foundation models [11,16] are typically fine-tuned for specific downstream diseases and tasks using multiple weakly supervised models [10,18]. However, their application in real-world scenarios is greatly hindered by the substantial training efforts required for various downstream tasks. In particular, self-supervised models[12,16,17] often involve substantial computational and expert resources to develop multiple classifiers for different tasks. Although some foundation methods[15,16] can bypass additional training, their performance is limited across foundation models and difficult to apply in clinical applications. This limitation primarily arises because pathological foundation models often rely on techniques such as MI-SimpleShot in UNI[16] and k -nearest neighbors (KNN) in HIPT[19]. While these general methods are designed to evaluate the effectiveness of foundation models across diverse tasks, they fail to account for the rich local information inherent in WSIs, resulting in suboptimal results. Moreover, zero-shot classification in vision–language foundation models[11,15] typically depends on millions of labeled pre-training data, while exhibiting limited effectiveness on specific tasks or hospitals outside the pretrained datasets. Thus, there is a pressing need to develop an effective approach using a few examples without training for generalizing foundation models across various downstream tasks in real-world hospitals, particularly in underserved regions that lack AI expertise and minority populations with limited cases.

In this paper, we introduce pan-cancer recognition without examples training (PRET), an approach that eliminates the need for task-specific model training. Inspired by the concept of ‘prêt-à-porter’ (abbreviated as pret, which refers to ready-to-wear clothing), PRET offers a ready-to-deploy solution for multi-cancer diagnostics and tasks. In addition to the advantage of training-free pan-cancer recognition, PRET fully leverages the rich local information from patch tiles for the effective use of limited labeled examples, beyond existing training-free methods[16,20,21]. Specifically, given a few labeled data as examples with a foundation model to extract their features, PRET learns from them during the inference phase in the manner of in-context learning (ICL)[22], without any parameter fine-tuning. Different from existing visual ICL methods for general images[23–26] or patch titles[27], PRET fully accounts for the unique characteristics of WSIs, where multiple instances (patch images) preserve rich local information, thereby facilitating exceptional recognition capabilities. In addition to its high effectiveness, PRET is versatile, supporting multiple task types, involving cancer screening, tumor metastasis detection, cancer subtyping and tumor segmentation in WSI analysis. Moreover, it accommodates various visual prompts including slide labels, bounding boxes, rough masks and tumor masks, which cater to a range of tasks for better performance and flexible labeling budgets.

To systematically assess the clinical utility of PRET, we conducted comprehensive evaluations across 23 benchmarks from multiple countries (the USA, China and the Netherlands), hospitals, organs, races and tasks. PRET consistently outperforms existing training-free approaches, achieving significant improvements in key performance metrics. For example, PRET exceeds the existing best-performing method by an area under the curve (AUC) of 36.76% ($P < 0.001$) at most and achieves high performances over 97% on 15 benchmarks. Notably, it attained clinical-grade accuracy in lymph node metastasis detection on the CAMELYON16 dataset [28], reaching an AUC of 98.71% ± 0.22% and surpassing the performance of 11 pathologists (average AUC of 81.0% ($P < 0.001$)) with just eight WSI examples. Additionally, PRET demonstrated robust generalizability across institutions and populations, with minimal performance variations (1.32% in non-small cell lung cancer (NSCLC) between racial groups and strong results in underserved regions (AUC = 98.72% ± 0.93%). These findings underscore PRET's potential to make advanced pathological AI systems more accessible, addressing critical disparities in cancer care worldwide.

## Results

### Pan-cancer recognition using a few examples without model fine-tuning

PRET introduces a paradigm shift in pan-cancer recognition by leveraging a few examples during the inference phase, completely eliminating the need for parameter fine-tuning. A single training-free PRET model demonstrated the capability to effectively generalize across a wide range of organs, datasets, hospitals and tasks, as shown in Fig. 1a. PRET eliminates the need for extensive labeled data, multiple models or parameter training in conventional[7,10], few-shot[16,29] and pan-cancer[11,15] solutions, as compared in Fig. 1b. This paradigm is highly flexible and practical, which can be readily applied to new tasks without AI experts to train models for individual tasks. This flexibility and low data cost represent excellent usability, reducing the cost of developing pathological AI models and better serving underserved regions lacking professionals and minority populations with limited cases.

PRET learns from a few examples with flexible visual prompts as Fig. 1c during the inference stage in the manner of ICL [22] without any parameter fine-tuning. As shown in Extended Data Fig. 1a, PRET consists of several proposed modules to achieve the above goal. Specifically, a pathological foundation model first extracts patch-level features, which are then processed by the in-context tagger that assigns diverse visual prompts (Supplementary Algorithms 1 and 2). Here, the tagged features are treated as visual in context, which is used to classify the discriminative feature instances from the test slide. The informative in-context classifier then identifies each test patch guided by these auto-tagged examples. Lastly, an attention-based aggregator combines patch-level results using self-attention on test features to generate a slide-level prediction, which is entirely parameter free. Detailed method comparisons can be found in Extended Data Fig. 1b–f.

### Single PRET model applied across diverse benchmarks

PRET is a versatile, training-free model designed for diverse pathological tasks and benchmarks. The involved tasks, datasets, organs and hospital divisions are summarized in Fig. 1a, with the data flow shown in Fig. 1d. PRET's architecture includes a feature extractor and in-context tagger that converts examples into visual in-context representations, processed by an in-context classifier for similarity-based matching with test instances. Rest modules (post-processor, instance miner and aggregator) are designed to support tumor segmentation, subtyping and slide-level prediction, respectively. We evaluated PRET on 23 benchmarks, covering 4,484 WSIs (~124 million patches) from 18 cancer types and nine organs across multiple international hospitals (Fig. 1a). Benchmarks included three task types and four prompt types, with 1–8 examples per task. Data details can be found in Extended Data Fig. 2a–d and Supplementary Table 1, including data scale, cancer subtypes, data split, tumor size and demographics.

PRET consistently outperformed existing methods [16,20,21], achieving superior performance in 20 of the 23 benchmarks (Fig. 1e) for their best performance. These baseline methods are used in foundation models without training, such as KNN-Mean in HIPT[19], MI-Prototype[20] and MI-SimpleShot[16], which rely on techniques[21] for general images. As compared in Extended Data Fig. 1b–d, these methods use fine-tuned classifiers, pooled slide-level features (KNN) or averaged dataset-level features (prototypes) to process examples. However, they lost rich local information of patch features within the slide, which is very useful in a limited data scale. In contrast, PRET processes all patch-level local features followed by an attention aggregator for the slide-level prediction, enabling superior performance on datasets such as CAMELYON16 (ref. [28]) and outperforming the best baseline by 36.76% ($P < 0.001$). With the elimination of parametric fine-tuning and corresponding overfitting, PRET can generalize well on CAMELYON17 (ref. [30]) for cross-hospital tests and scan corruptions in CAMELYON16-C (ref. [31]), with AUC improvements beyond the best-performing baseline method (MI-SimpleShot [16]) by 17.84% ($P < 0.001$,89.42% ± 0.81% versus 71.58% ± 2.5%) and 32.87% ($P < 0.001$,95.27% ± 0.76% versus 62.40% ± 2.75%), respectively. PRET consistently delivered high performances across diverse tasks, achieving AUCs exceeding 97% on 15 benchmarks, such as colorectal cancer (CRC) screening (100%), renal cell carcinoma (RCC) subtyping (97.07% ± 1.03%) and esophageal squamous cell carcinoma (ESCC) tumor segmentation (99.54% ± 0.15%). These results highlight the method's effectiveness across various tasks.

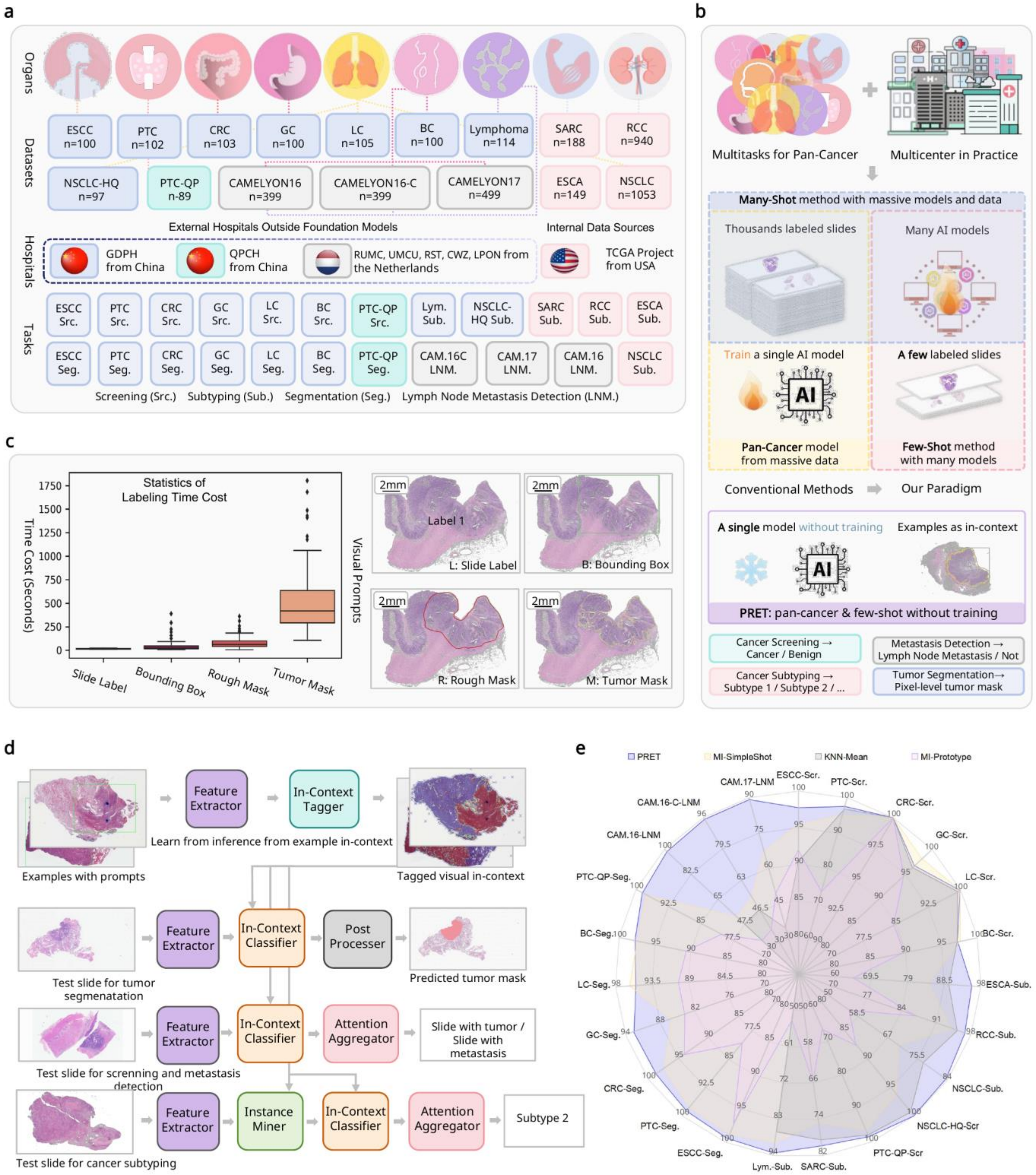

**Figure 1. Overview**. **a**, PRET supports multiple organs, datasets, hospitals, tasks and label types (visual prompts). We propose multiple datasets from GDPH and QPCH and add various prompts for the involved datasets. Hospitals outside of TCGA are external medical centers that do not participate in the pre-training or fine-tuning processes. **b**, Comparison of PRET to existing solutions. PRET eliminates the need for massive labeled data or multiple models. The fire indicates model fine-tuning and the snow indicates no training. **c**, Labeling time cost, counted from 79 random slides from multiple organs and 11 pathologists. Examples of slides and visual prompts. The slide label is a certain number, while other visual prompts are boxes or outlines given in these examples. **d**, The data flow of PRET is to be applied to multiple tasks. **e**, PRET surpasses training-free methods of foundation models in large-scale evaluations, involving 23 tasks about cancer screening and subtyping, tumor segmentation and metastasis detection at their best AUC.

**PRET leverages flexible visual prompts for accurate cancer screening and subtyping**

We introduce a range of flexible visual prompt types to facilitate data labeling across various tasks and labeling budgets. These prompts include instance-level tumor masks (M), as well as weak prompts such as slide labels (L), bounding boxes (B) and rough masks (R). Figure 1c illustrates examples of these prompts, with bounding boxes and rough masks being contributions to pathological recognition. These prompts are incorporated into newly constructed datasets and we supplemented The Cancer Genome Atlas (TCGA) datasets with B, R and M for prompts. Labeling efficiency varied, with slide labels requiring an average of 15.4 s, while B, R and M prompts took 44.5 s, 85.1 s and 537.2 s, respectively (Fig. 1c). For PRET, the local visual in-context tagger module is used to dynamically tag each instance with weak prompts for subsequent recognition.

We extensively evaluated the proposed PRET on slide-level recognition tasks, including cancer screening and subtyping. As shown in Fig. 2a, the first row represents the screening task results, while the second row shows the subtyping outcomes. PRET outperformed baseline methods across all datasets, task types and varying prompts and example quantities. In cancer screening, PRET achieved high AUC results exceeding 95% on ESCC, CRC, lung cancer (LC) and breast cancer (BC) benchmarks for the average of 1–8 examples across prompts, with the highest mean AUC of 99.96% for CRC. Even with only a single example, PRET maintained strong performance, with AUC values of 96.27%, 99.97% and 98.04% on ESCC, CRC and LC benchmarks, respectively. Moreover, our model significantly outperformed the best-performing method by margins of 8.24% ($P = 0.008$) for ESCC for with a single slide label (one-shot), 11.03% ($P = 0.002$) for papillary thyroid carcinoma (PTC) two-shot using slide labels in Supplementary Table 2 and 5.99% ($P = 0.003$) for gastric cancer (GC) one-shot with tumor mask prompt. Subtyping tasks present greater challenges than cancer screening but PRET continued to deliver substantial improvements, significantly surpassing the best-performing method by 9.03% for lymphoma ($P < 0.001$, four-shot AUC at 85.62% ± 7.9% with box prompt), 15.66% for esophageal cancer (ESCA) subtyping ($P < 0.001$, four-shot AUC at 91.71% ± 2.69% with slide label) and 14.41% for NSCLC ($P < 0.001$, two-shot AUC at 82.58% ± 4.12% with slide label). Moreover, we compared our method using more evaluation metrics such as F1 score and accuracy (Supplementary Tables 2–9), where our method was superior in almost each result. Detailed comparisons are provided in Extended Data Fig. 3, demonstrating the broad and great effectiveness of our approach.

We analyzed the optimal prompts and example quantities for each benchmark and their corresponding performances (Fig. 2b). Results showed that different tasks favor different prompt types, offering guidance on selecting the most suitable prompt and number of examples. We further examined the relationship between example quantity and performance, finding that more examples generally improve results, although high-quality individual examples could sometimes outperform larger sets (Fig. 2c). The distributions of L, M and B + R prompts were roughly one-third each (Fig. 2b). According to these results and tumor size statistics (Extended Data Fig. 2c), smaller tumors (for example, PTC and ESCC) benefited most from M, B, and R prompts, with M yielding the best accuracy and B/R offering lower annotation costs. In contrast, larger tumors (for example, CRC, lymphoma and NSCLC) were easier to localize and performed well with L prompts, which require less positional detail. On the basis of this analysis, we provide three prompt selection guidelines. Firstly, slide labels from diagnosis records are labeling free and optimal in about one-third of tasks (Fig. 2c). Secondly, if slide labels perform poorly, users can refer to Fig. 2b to select prompts for similar cancer types, where smaller tumors generally benefit from more detailed prompts. Thirdly, determining the optimal prompt under the one-shot setting and then increasing the number of examples is an efficient strategy for improvement. In addition to prompt-specific analysis, PRET consistently outperformed existing methods across multiple datasets for all four prompts with varying shots of examples ($P < 0.002$), as shown in Fig. 2d.

It is worth mentioning that our method is ideal for pathological foundation models and complements them. Motivated by stronger pathological representations from foundation models, we realized that pan-cancer recognition could break free from fine-tuning numerous downstream models, adopting a training-free approach that saves substantial data and model training efforts. As shown in Fig. 2e, the conventional non-pathological model (ResNet-50 (ref. [32]) pretrained from ImageNet[33]) limited the effectiveness of PRET because of weak representations. With the emergence of pathological foundation models, PRET has greatly surpassed other methods, enhancing the performance of foundation models to recognize cancers without training. For example, the performance of NSCLC subtyping for the Prov-Gigapath model[13] improved from 73.99% AUC to 90.18% in eight-shot and these improvements were significant across a wide range of pathological foundation models[11–13,16,17,34–36].

Beyond quantitative metrics, qualitative visualizations in Fig. 2f revealed that PRET could successfully locate the tumor region using a weak visual prompt. By leveraging the visual in context of these tagged features, our method generated high responses in positive regions and minimal responses in negative samples (Fig. 2g), highlighting its clear discriminative ability. Furthermore, we provide visual comparisons to baseline methods, further illustrating the effectiveness of our approach (Extended Data Fig. 4).

**PRET outperforms fine-tuned methods in few-shot setting**

In addition to the comparison to training-free methods in a fair setting, results in Fig. 3a suggest that PRET greatly outperformed conventional task-specific fine-tuning methods [10,18,37] (for example, AUC of 82.58% ± 3.98% for PRET versus 71.64% ± 6.86% for the best baseline TransMIL[37] on NSCLC two-shot setting), even if they conducted an extra training phase. Specifically, PRET retained rich local information from all patch features, which allowed for more effective use of data under a limited scale. Additionally, our method does not require parameter fine-tuning, which helps avoid parametric overfitting under limited samples. In contrast, the parameters of the attention modules in conventional methods are difficult to optimize when the data scale is insufficient, resulting in performance that is far inferior to our method. More method comparisons and analyses are provided in Extended Data Fig. 1.

We also compared fine-tuned methods using a wide range of foundation models in Fig. 3b. Results suggest that PRET consistently surpassed fine-tuned methods for 1–8 shots, with wide applicability to foundation models. For example, PRET improved fine-tuned TITAN[36] from an AUC of 61.41% ± 6.51% to 95.15% ± 1.32% at a gain of 33.74% for one-shot RCC subtyping. PRET is more robust in a limited data scale without parameter training, whereas fine-tuned foundation models rely on more parameters and their self-attention modules are hard to optimize in the one-shot setting. When scaling update data size to eight WSIs per class, PRET still greatly outperformed fine-tuned foundation models on the NSCLC subtyping task, with an AUC of 89.91% ± 5.27% in contrast to 84.41% ± 13.08% for TITAN [36],67.73% ± 12.37% for CHIEF[12],73.41% ± 14.17% for Pro-GigaPath[13],77.31% ± 13.79% for VirChow[34] and 74.86% ± 14.98% for UNI[16]. More result comparisons are available in Extended Data Fig. 5a, with hyperparameter analysis in Extended Data Fig. 5b,c.

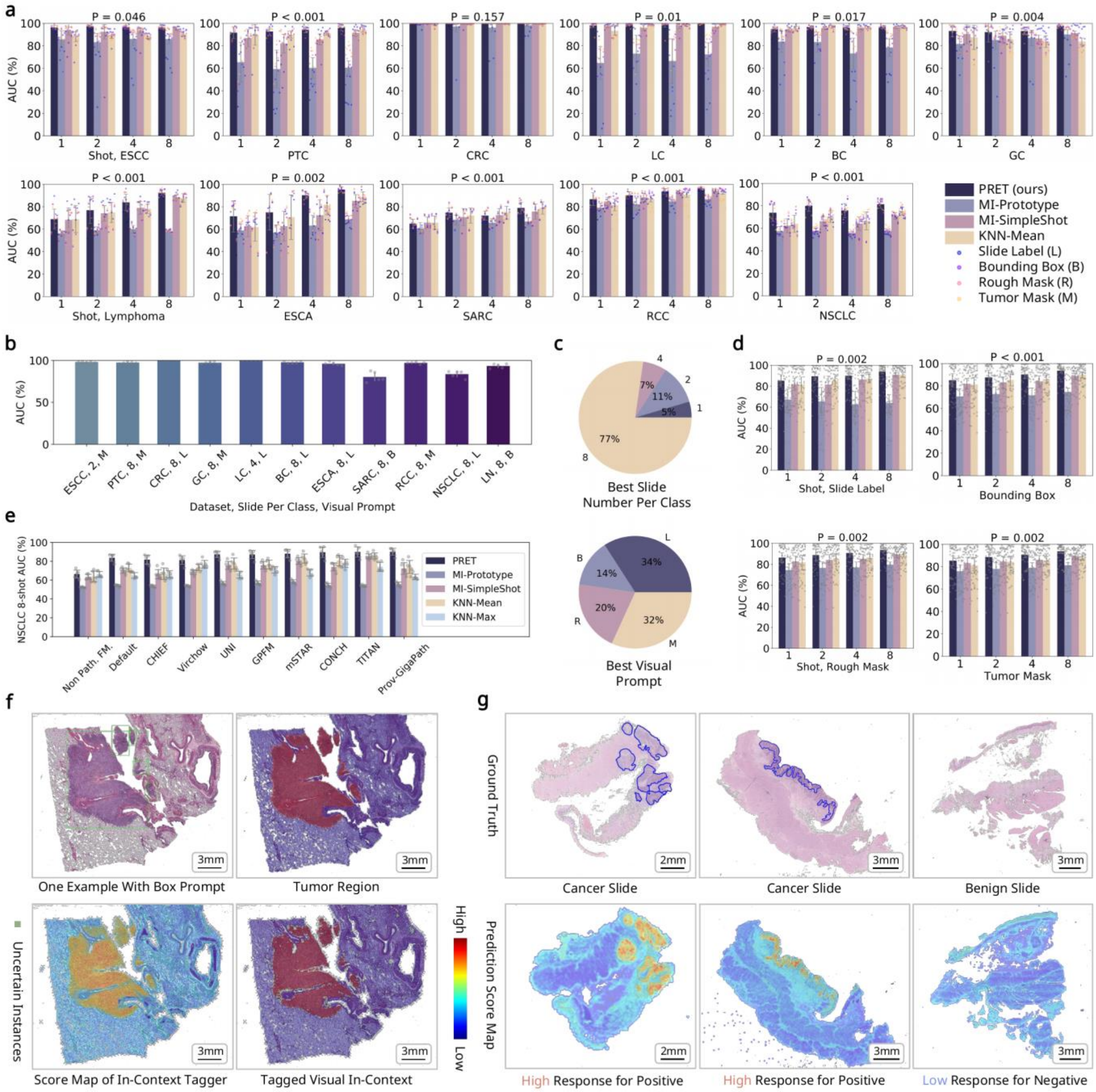

**Figure 2. Performances of cancer screening and subtyping**. **a**, Pan-cancer performances of cancer screening and subtyping tasks. Our method showed higher results than baselines in all involved datasets and example numbers (shot). **b**, Best results for each dataset, slide numbers and prompt types, suggesting that different prompts flexibly help to enhance the performance. **c**, Statistics about best slide numbers and best visual prompts. More examples are generally better and the proportions of L, M and B + R were close. **d**, PRET performed best at any prompt on average. **e**, PRET is widely complementary to pathological foundation models and beyond other training-free methods. Notably, PRET is ideal for foundation models with much larger performance gaps compared to nonfoundation models (ResNet-50 (ref. 32) pretrained from ImageNet). **f**, The in-context tagger generates tagged visual in-contexts that are close to the manually labeled tumor region on the NSCLC dataset. **g**, Our method is visually distinguishable between cancer slides and the benign slide on the ESCC dataset with high explainability. The error bars in **a**, **b**, **d**, **e** show the s.d. across multiple independent experiments, where n = 20 for a (four prompts and five repeats), n = 5 for **b**, **e** (five repeats) and n = 55 for d (11 benchmarks and five repeats). The P values in **a**, **d** indicate the significance of PRET outperforming the best baseline according to a two-sided Wilcoxon test, reporting the median value across all repeat experiments and varied shots.

## PRET enables tumor segmentation from weak labels

In addition to classification tasks, PRET facilitates accurate tumor segmentation from weak prompts. As shown in Fig. 3c, our method outperformed the baselines across different datasets. Specifically, PRET achieved Dice improvements beyond the best-performing method (MI-SimpleShot[16]) by 17.38% (P < 0.001, from an AUC of 69.24% ± 24.29% to 86.62% ± 4.99%) and 14.61% (P < 0.001, from an AUC of 63.21% ± 18.97% to 77.82% ± 7.40%) for ESCC and PTC, respectively, with only eight slide labels (Supplementary Table 6). We also compared the results of weak prompts, L, B and R, against the strong prompt M (Fig. 3d). Overall, PRET demonstrated smaller discrepancies between strong and weak prompts, with differences typically under 5% (even below 1% for BC). In contrast, methods such as MI-Prototype and MI-SimpleShot exhibited gaps exceeding 20% and 10%, respectively. This suggests that PRET is more adept at handling weak prompts, reducing annotation costs while ensuring model performance. Further comparisons revealed many results exceeding 90% (Extended Data Fig. 5d and Supplementary Tables 6–9), demonstrating the high effectiveness of our method.

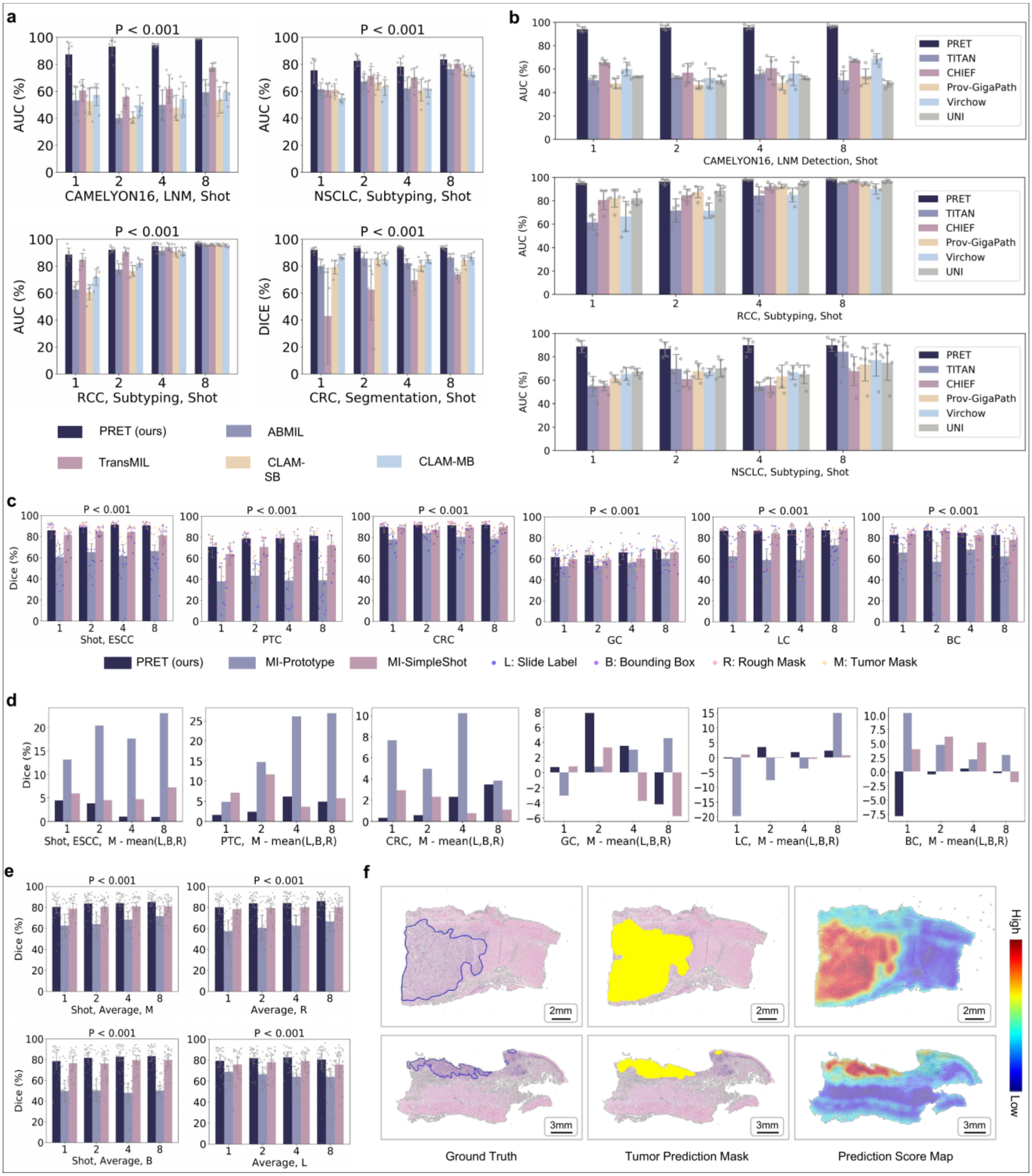

**Figure 3. Performance** in **tumor segmentation. a**, PRET eliminates the need to fine-tune parameters, beyond conventionally weakly supervised methods (ABMIL[18], TransMIL[37], CLAM-SB[10] and CLAM-MB[10]) in performance and development cost, using the default foundation model[17]. The used prompt types correspond to Fig. 2b. **b**, PRET outperforms a wide range of fine-tuned foundation models. These foundation models were fine-tuned using their own slide-level classifier; otherwise, CLAM-MB[56] was applied for fine-tuning. PRET uses patch features of TITAN[36] and all data splits were the same for fair comparisons. **c**, PRET outperforms training-free methods of foundation models in almost all the datasets, shots and prompts. **d**, Weak prompts (L, B and R) on the proposed PRET show closer results to the strong tumor mask prompt than other methods, indicating better weakly supervised segmentation ability. Fluctuations sometimes result from the quality of the examples in a few shots. **e**, PRET performs best at all prompts on average of multiple datasets. **f**,

Visualization of segmentation results, including the predicted mask and score map. The error bars in **a** – **c, e** indicate the s.d. across multiple independent experiments, where n = 5 for **a**, **b** (five repeats), n = 20 for **c** (four prompts with five repeats) and n = 30 for **e** (six benchmarks and five repeats). The P values (a**, c, e**) indicate the significance of PRET outperforming the best baseline according to a two-sided Wilcoxon test, reporting the median value across all repeat experiments and varied shots.

Interestingly, we observed that weak prompts occasionally outperformed tumor masks, such as in the case with the BC's one-shot example. This could be because of tumor masks annotated by doctors occasionally including excessive tumor stroma or other non-cancer tissues, whereas our algorithm mitigates these interferences using the in-context tagger. Further evidence is given in Fig. 3e, where the rough mask presented higher performance gaps beyond other methods without the in-context tagger. Additionally, qualitative comparison to baselines showed that our method predicted tumors more accurately and continuously, with fewer obvious errors occurring in MI-Prototype, demonstrating better stability under weak prompts (Extended Data Fig. 6). Visual results in Fig. 3f further confirm that our method closely aligned with manually annotated tumor masks, highlighting its capability to generate accurate tumor masks from weak prompts. Moreover, these segmentation results can be applied to enhance recognition tasks because the in-context tagger relies on discriminative instances, as illustrated in Extended Data Fig. 7.

## PRET achieves clinical-grade performance in lymph node metastasis detection with strong generalizability

Lymph node metastasis detection poses a challenging task for pathologists, as metastatic tumors are often very small and easily overlooked. PRET addresses this challenge by preserving rich local information and ensuring the inclusion of even tiny tumors. On the CAMELYON16 (ref. [28]) benchmark, when the specificity was 0.9875 (1 − specificity = 0.0125), our method outperformed the reported pathologist performances in sensitivity by 42.86%, 36.74%, 26.53% and 16.33% in their routine diagnosis time, achieving the optimal sensitivity of 0.9388 (reaching the optimal sensitivity of pathologist without time constraints) (Fig. 4a). Across repeated experiments, PRET maintained AUCs of 0.9834–0.9901 (average: 0.9871% ± 0.22%), surpassing the pathologists' average AUC of 0.810 (0.738–0.884), $P < 0.001$. These results show that PRET delivers stable, clinical-grade performance in lymph node metastasis detection.

In addition to outperforming pathologists, our algorithm surpassed existing few-shot methods [20,21,29] and even outperformed many-shot approaches [18,37–40] that require more data and model fine-tuning. Our algorithm greatly outperformed existing non-parametric methods including MI-Prototype [20], MI-SimpleShot [16] and KNN, as well as weakly supervised methods with extra training such as ABMIL [18], CLAM-SB [10] and UNI + ABMIL [10] and prompt tuning-based TOP [29] (Fig. 4b). This indicates that the effectiveness of PRET exceeds other types of few-shot methods. In comparison to many-shot methods, PRET achieved superior performance using only eight examples, whereas conventional methods used 243 WSIs. Despite reducing our dataset size by approximately 30-fold, PRET still outperformed state-of-the-art algorithms, such as TransMIL [37], CLAM[10], ABMIL [18], TransMIL + FT[38], MHIM-MIL [39], UNI + ABMIL[16] and dMIL-Trans[40] (Fig. 4c). These results indicate that our training-free approach offers notable advantages over conventional fine-tuning methods in terms of performance and development cost, with high effectiveness and practicality.

A key advantage of PRET is that it does not require training, offering greater flexibility and contributing to strong generalization capabilities. By avoiding possible overfitting during the parameter fine-tuning process, PRET can effectively adapt to different hospitals and scanners using dynamic similarity computation for each slide. As shown in Fig. 4d, PRET demonstrated remarkable stability under challenging scanning conditions such as stains and out-of-focus images in the CAMELYON16-C (ref. [31]) benchmark. When directly applying models obtained from CAMELYON[16] to test on CAMELYON16-C, the performance gap was only 3.44% ($P < 0.001$), significantly outperforming the best existing method by 7.55% ($P < 0.001$). Furthermore, when tested on the multicenter CAMELYON17 dataset in Fig. 4e, PRET consistently achieved superior performance beyond all baselines.

To better understand the source of PRET's improvements, we conducted a qualitative comparison to show that PRET can accurately locate tumor regions for positive samples (Fig. 4f). Notably, PRET exhibited minimal response for negative slides without metastasis, in contrast to MI-Prototype and MI-SimpleShot, which exhibited false positive regions. This indicates that PRET can better differentiate between positives and negatives, contributing to its higher recognition performance.

**PRET can effectively serve minority populations, underserved regions and external hospitals**

Developing pan-cancer recognition AI systems typically requires extensive labeled data and costly training resources, which limits accessibility in data-scarce regions. PRET addresses this challenge by removing the dependence on large datasets, enabling affordable, rapid development of pathological AI systems in resource-limited settings. It also shows strong generalization capability, maintaining stable performance when applied to external hospitals.

To assess generalization across populations, we evaluated PRET on the NSCLC benchmark from TCGA (Fig. 5a), covering white, black and Asian or unreported (A&U) groups (A&U merged for sufficient data). Using slide labels from white and black subgroups (Fig. 5b), PRET showed minimal racial disparity, with an eight-shot example setting showing only 1.32% maximum difference across races, far lower than other methods. In the one-shot setting, racial performance gaps remained below 10%, confirming PRET's robustness in low-data scenarios. Importantly, the overall performances using examples from different racial groups were relatively consistent, underscoring PRET's robust generalization across populations (Fig. 5b).

To evaluate PRET's applicability in an underserved region, we assessed its performance on the dataset from Qinghai Province, a region ranked 33rd of China's 34 first-level administrative divisions by GDP. Qinghai's lower economic status and sparse population resulted in limited data availability that posed challenges for developing complex models that require extensive data. Herein, PRET showed outstanding performances in cancer screening and tumor segmentation on the PTC benchmark of Qinghai Province (PTC-QP), with superior performance over the best-performing existing method by 9.85% AUC ($P = 0.005$) on screening and 8.33% Dice ($P < 0.001$) on segmentation using just one example (Fig. 5c). It achieved AUCs of 98.72% ± 0.93% for screening and 98.82% ± 0.14% for instance level, as well as a Dice score of 89.85% ± 0.75% for segmentation.

Moreover, compared to leading zero-shot foundation models (PLIP[15], CONCH[11] and MUSK[41]), PRET showed significant improvements (Fig. 5d). On CAMELYON16, PLIP[15] and CONCH[11] achieved ~60% accuracy, while PRET reached 95.35% ± 0.49% accuracy and 98.71% ± 0.22% AUC, a 21.29% gain ($P < 0.001$) over the best prior method (MUSK[41]; 74.06% accuracy). PRET continued to improve with more examples, achieving 98.15% ± 1.55% on NSCLC-HQ and 94.04% ± 3.40% on PTC-QP, surpassing all prior zero-shot foundation models.

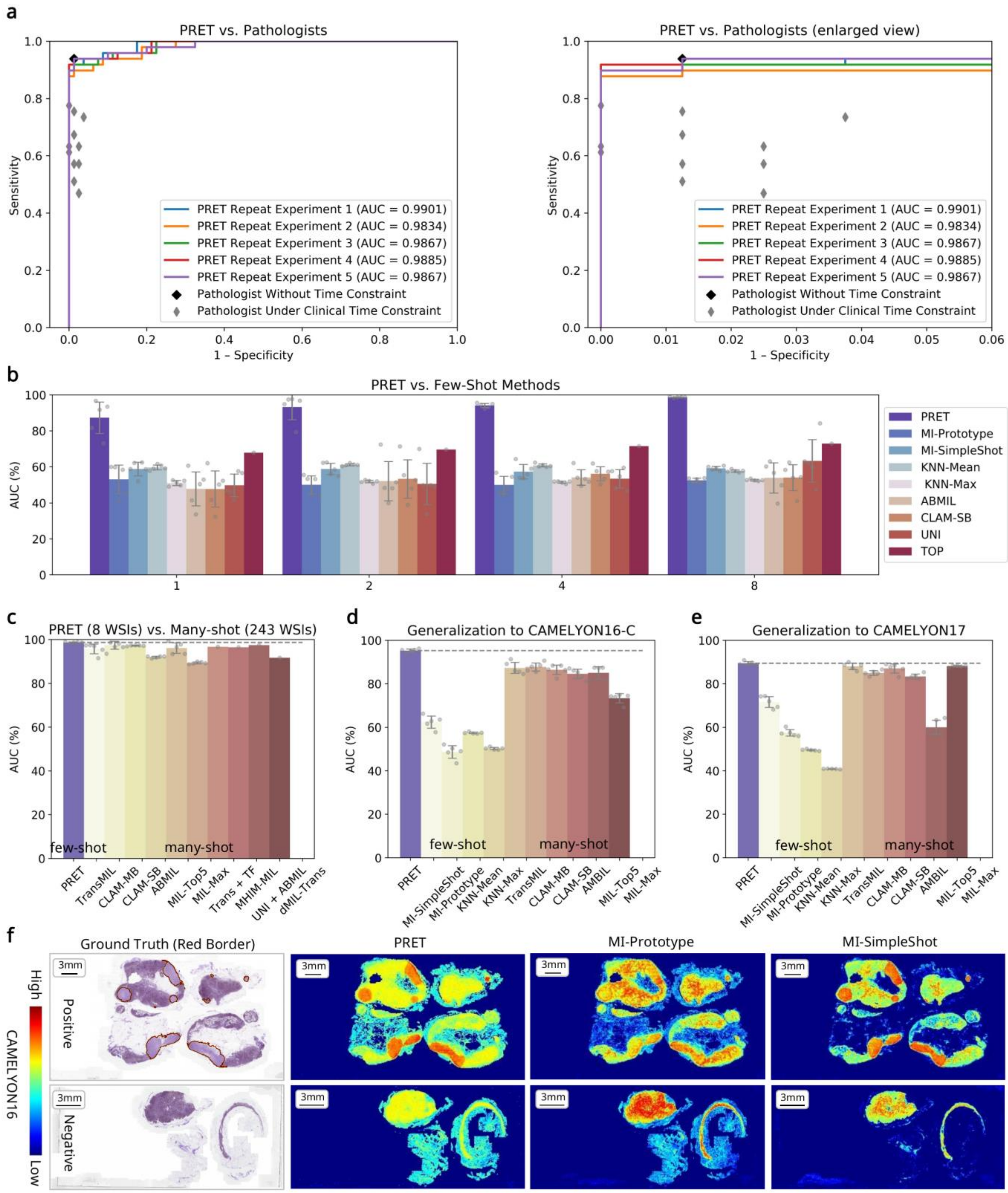

**Figure 4. Clinical-grade performances on lymph node metastasis detection with strong generalization. a**, PRET versus pathologists. The ROC curves of PRET over scatters indicate that the proposed method presents higher sensitivity than pathologists under the same specificity. The performances of pathologists are reported in CAMELYON16 (ref. [28]); the gray scatters denote the detection of metastasis under clinical time constraints. Right: subplot providing a cropped and enlarged view for enhanced clarity in comparison. **b**, Comparison of PRET to few-shot learning methods. **c**, PRET using eight slides even surpasses many-shot methods that use 243 slides on the CAMELYON16 dataset. **d**, **e**, PRET shows higher generalization than both few-shot and many-shot methods for slides with scan corruption (d; CAMELYON16-C (ref.

[31])) and multiple outer centers (e; CAMELYON17 (ref. [30])). **f**. PRET is more distinguishable between positives and negatives. The error bars in b – e indicate the s.d. across n = 5 independent experiments. Results without error bars (for example, TOP[29] and TransMIL + FT[38]) are reported from their papers without repeats.

We explored the generalization ability on external hospitals, as shown in Fig. 5e, with performance comparisons across specific hospitals. The first lymph node metastasis detection task involved two internal hospitals (University Medical Center Utrecht (UMCU) and Radboud University Medical Center (RUMC)) from the CAMELYON16 dataset[28] and three extra external hospitals (Canisius-Wilhelmina Hospital (CWZ), Rijnstate Hospital (RST) and LabPON (LPON)) from the CAMELYON17 dataset[30] for slides with lymph node metastasis at micro and macro sizes. It was observed that PRET achieved similar results between internal (RUMC = 95.77% AUC) and external hospitals (for example, CWZ = 95.15% AUC) under one-shot settings. For the second PTC screening task, AUC differences only ranged from 1.01% (eight-shot) to 2.55% (two-shot) by comparing labeled examples from the Guangdong Provincial People's Hospital (GDPH) and Qinghai Province Cancer Hospital (QPCH) centers. For the third NSCLC subtyping task, we conducted external experiments by applying examples from NSCLC-TCGA to NSCLC-HQ (from the GDPH center). The results were close, with AUC scores of 98.54% versus 98.27% for different sources of labeled examples in the two-shot setting. The lower results occurred for the internal setting of NSCLC (TCGA) instead of the generalized external hospital. Figure 5f reveals that high-quality slides could improve the performance of PRET, whereas external examples were robust during cross-hospital generalization, as discussed above.

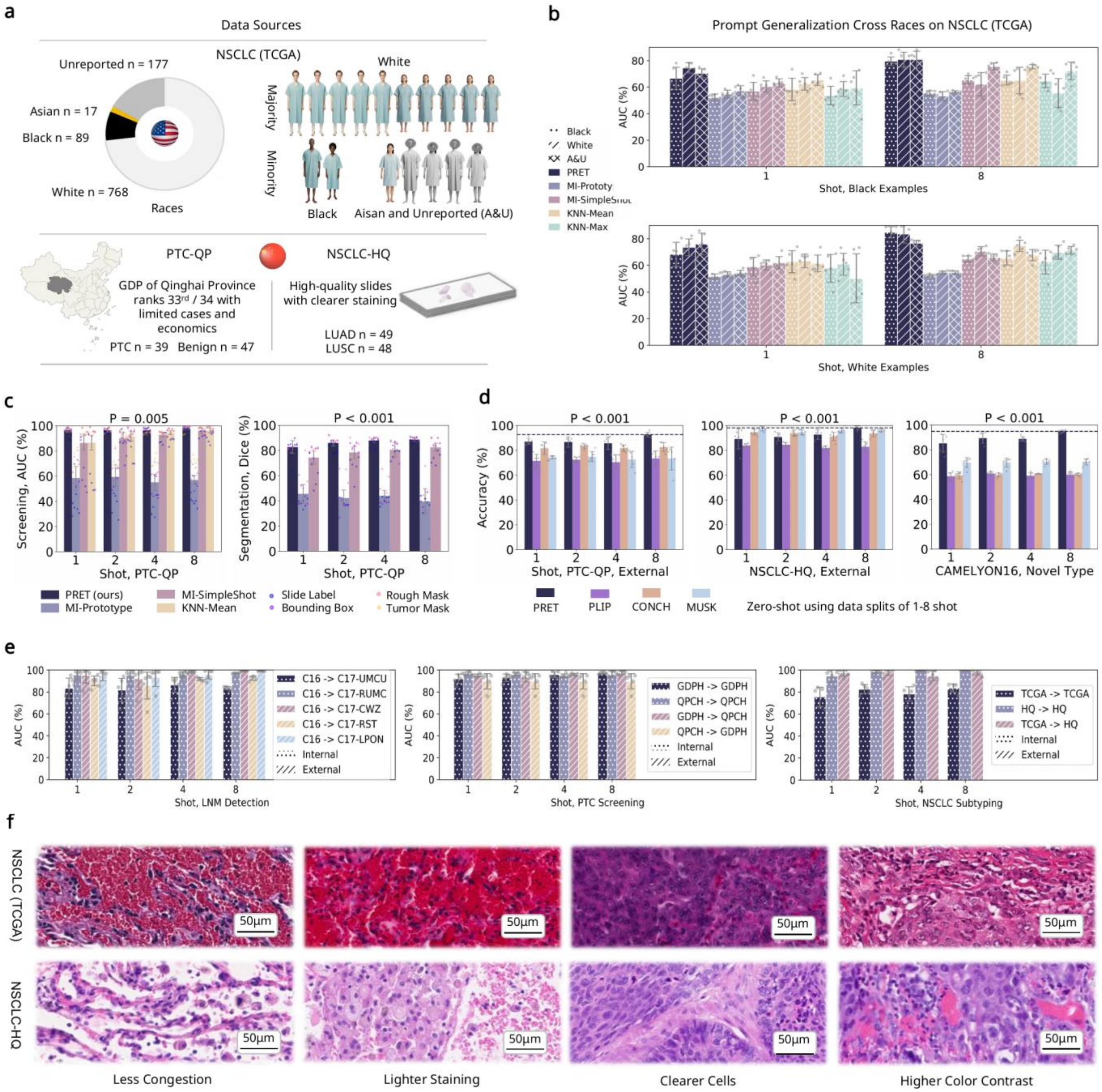

**Figure 5. Generalization performances on minority populations, underserved regions and external hospitals.** PRET can readily serve minority populations and underserved regions, with strong generalization across hospitals. The adopted prompt type in this figure corresponds to Fig. 2b. **a**, Illustration of data statistics. In TCGA NSCLC, black and Asian groups are the minority. The Asian groups with the least cases were merged with unrepeated races (A&U) to ensure sufficient data scale. PTC-QP represents underserved regions, where the GDP ranks 33*rd* of 34 in the first-level administrative regions of China. NSCLC-HQ is a high-quality benchmark, where LUAD indicates the lung adenocarcinoma and LUSC indicates the lung squamous cell carcinoma. **b**, Performance comparison among races. Two example types (black and white) are applied to three racial groups to assess cross-racial generalization. PRET only shows a 1.32% AUC gap across races for eight-shot black examples. **c**, PRET achieves high performances on PTC-QP for both cancer screening (P = 0.005) and tumor segmentation (P < 0.001) tasks. The P values indicate the significance of PRET outperforming the best baseline according to the two-sided Wilcoxon test at a median value across n experiments and varied shots. **d**, PRET can readily support external hospitals with higher results (P < 0.001) than the zero-shot foundation models PLIP[15], CONCH[11] and MUSK[41], which are trained from massive textual labels. The dashed lines are the best results for each method. Methods apply the same data splits under varied shots for fair comparisons. The P value is the median value across experiments according to the two-sided Wilcoxon test. **e**, PRET presents strong generalization on external hospitals across diverse tasks. ‘ A → B’ indicates testing on B center using labeled examples from A. The involved hospitals

include UMCU, RUMC, CWZ, RST, LPON, GDPH, QPCH, TCGA and HQ (GDPH), where bars with slashes are external experiments. C16 indicates the CAEMLYON16 dataset, containing data from UMCU and RUMC. **f**, Case comparisons between NSCLC-HQ and NSCLC (TCGA) including benign and tumor regions. Pathologists can readily improve PRET using more examples or by increasing the slide quality, including less congestion, lighter staining and other factors leading to clearer slides. The error bars in b – e indicate the s.d. across multiple independent experiments, where n = 5 for **b**, **d**, **e** (five repeats) and n = 20 for **c** (five repeats and four prompts).

Overall, PRET delivers superior generalization, adaptability and practicality, providing an effective, scalable solution for AI-driven cancer diagnostics across diverse and resource-constrained settings.

## Discussion

Our proposed PRET enables pan-cancer recognition with only a few examples, allowing a single model to rapidly recognize multiple cancers and perform diverse tasks across organs, hospitals, races and regions without additional training. We evaluated PRET on 23 benchmarks, comprising 4,484 WSIs (124 million patches) from multiple international hospitals, including underserved regions and underrepresented racial groups, underscoring its universality and societal impact. PRET outperformed existing training-free methods of foundation models [16,19] on 20 tasks and achieved clinical-grade diagnostic performance for lymph node metastasis detection, surpassing pathologists. Compared to conventional methods [7,18], PRET offers several key advantages: (1) low data cost, with no need for large labeled datasets; (2) single pan-cancer model, with consolidation of tasks into one universal model; (3) training-free usability, with no retraining for new tasks; (4) high generalization, with robust performance across populations and hospitals; and (5) flexibility, allowing easy integration with existing foundation models and dynamic adaptation using ICL. PRET excels in tasks requiring local information, such as detecting lymph node metastases. Its non-parametric classifier prevents overfitting by adapting dynamically to each test slide, enabling exceptional cross-hospital and cross-racial generalization. Furthermore, integration with ICL further enhances the inference by leveraging pretrained representations from pathological foundation models. Despite its success, PRET has limitations. Firstly, retaining local information in the classifier increases inference time (7.41 s per WSI under eight examples on the CAMELYON16 (ref. [20]) test set), although this is negligible compared to day-long slide preparation. Secondly, inter-hospital variability persists (for example, eight-shot performance gaps of 9.29% and 16.34% for lymph node metastasis (Fig. 4e) and NSCLC (Fig. 5e), versus smaller gaps for racial differences (1.32–8.2%) and scan failures (3.44%)). As illustrated in Fig. 5f, staining inconsistencies (for example, darker TCGA slides versus cleaner GDPH slides) were a major cause of this phenomenon, which could be mitigated by improving slide quality. Thirdly, reliance on few examples can cause incidental errors when unique tissue patterns appear only once. While increasing the number of examples can mitigate this issue, the inherent randomness associated with small example sizes remains a challenge. Lastly, PRET can scale performance with more examples (Extended Data Fig. 8a), whereas fine-tuned models with much more data may exceed PRET's performance on subtyping tasks (Extended Data Fig. 8b). Nevertheless, PRET achieved comparable or superior performance with far fewer samples, demonstrating its high potential and practical value.

Although current results are promising, growth opportunities are abundant. Firstly, improving the quality of testing slides and example data could further elevate model performance, potentially enabling more tasks to reach clinical-grade accuracy. Additionally, as demonstrated in Fig. 2e, using stronger pathological foundation models in practical applications is expected to enhance performance over time. Secondly, task expansion represents another promising avenue. Beyond classification, PRET has potential applications in tasks such as mutation prediction[42] and regression-based tasks such as prognosis[43], which could drive further algorithmic innovations. Thirdly, integrating PRET's visual ICL capabilities with emerging vision–language models[44] or multi-modal large language models[45] could unlock further possibilities. Innovations in local ICL and the use of diverse visual prompt types may serve as heuristic insights for future research, paving the way for advancements in pathological AI.

## Methods

### Pan-cancer recognition using examples without training

**Preliminaries and overall framework.** In this work, we introduce PRET, an innovative framework that enables a single foundation model to perform multiple tasks by learning from a few examples during the inference phase (Fig. 1b). By shifting from traditional textual examples to labeled image examples, we apply ICL to WSI recognition, enabling flexible task execution with minimal prior training. PRET supports tasks such as cancer screening, segmentation and subtyping through modular components, which can be combined as needed for different applications, summarized in Supplementary Table 10. The PRET framework consists of six main modules: (1) feature extractor, where test WSIs and example instances are encoded into feature representations using a pathological foundation model; (2) in-context tagger, where, for the example feature instances, the in-context tagger assigns labels from various visual prompts, providing feature-level visual in context for different tasks; (3) discriminative instance miner, which identifies tumor regions necessary for subsequent subtyping tasks (for screening and segmentation, the entire WSI is processed, while subtyping only requires the classification of tumor regions); (4) informative in-context classifier, which calculates similarities between test and example instances, enabling non-parametric classification for each test instance; (5) attention aggregator, where, for tasks requiring slide-level representation, such as screening and subtyping, the attention aggregator performs a weighted sum of relevant instances to produce a global score; (6) post-processor, where, for segmentation tasks, Gaussian smoothing is applied to produce continuous pixel-level predictions. Note that PRET was developed from our previous work INC[46], where its non-parametric classifier is applied to realize the informative in-context classifier. The difference is that we expanded the previous lymphatic metastasis classification to pan-cancer recognition through ICL, supporting various prompts using the in-context tagger module instead of the previous tumor mask only and further supporting subtyping and segmentation tasks through the miner and post-processor modules. Most importantly, we conducted large-scale pan-cancer experiments, clinical comparisons, cross-racial comparisons and regional experiments, systematically demonstrating the importance of PRET. The modules involved in the overall process are described in Supplementary Algorithm.

### Implementation of PRET

**Feature extractor.** We use the pretrained model proposed by Kang *et al.*[17] as the default feature extractor. This foundation model was pretrained on pathological images from TCGA without any label in a self-supervised manner on the basis of the DINO [47] framework in the ViT-Small [48] architecture with a patch size of 8. Moreover, we applied stronger pathological foundation models in Fig. 2e and Extended Data Fig. 5a to demonstrate the extensibility and compatibility of PRET including CHIEF[12], Virchow[34], UNI[16], GPFM[35], mSTAR[49], CONCH[11], TITAN[36], Prov-Gigapath[13] and MUSK[41]. Specifically, CHIEF[12] applies CTransPath[50] as the patch-level feature extractor and trains the slide-level feature using contrastive learning. Virchow[34] was pretrained from approximately 1.5 million hematoxylin and eosin (H&E)-stained WSIs acquired from Memorial Sloan Kettering Cancer Center using ViT[48] and DINOv2 (ref. [51]). UNI[16] was pretrained on their unlabeled Mass-100K dataset with the DINOv2 (ref. [51]) using ViT-Large [48] at a patch size of 16 with a larger data scale and network size. GPFM was trained from 33 public datasets using ViT-L[48] with the proposed UDK method.

CONCH[11] is supervised by text using over 1 million image–text pairs using conservative learning[52] based on ViT-Base[48], whose image encoder is used to extract features. TITAN[36] was pretrained using visual self-supervised learning and vision–language alignment with CONCH (version 1.5)[11], with focus on multi-modal slide-level representations. Prov-Gigapath[13] uses 171,189 whole slides from Providence pretrained with DINOv2 and LongNet architecture[53]. For each foundation model, we used the same strategy to extract features for a fair comparison, except TITAN[36], whose input size is 512 ×512. Specifically, we sliced each WSI into multiple patches with a size of 256 at ×20 magnification. Patch images larger than 5 kB were considered foreground for the subsequent feature extraction step. The size limit was set to 15 kB for the slide label to exclude inferences such as stains because of their weaker constraint compared to other visual prompts. For the valid patches, we applied five crops with a size of 224 to fit the network input size and extract features without other augmentation. Then, L2 normalization was applied to the mean feature of crops and we obtained the feature instances expressed in Eq. 1 in Supplementary Algorithm.

**In-context tagger.** As shown in Fig. 1d, the in-context tagger, along with the feature extractor, was common for each task. The tagger is responsible for processing the visual prompts for the next recognition steps. For this work, we used one, two, four and eight examples with prompts in this work. The instance label of the tumor mask prompt was used directly and other prompts were tagged in varying situations according to the tasks, as described in Supplementary Information. The involved hyperparameters include the uncertain range factor $t_e$ to divide uncertain instances, which was grid-searched on the validation set (Extended Data Fig. 5b). If the validation set is not available, the default value (0 for $t_e$) and that searched on other available benchmarks performed closely (Extended Data Fig. 5c), demonstrating high robustness of the hyperparameters.

**Cancer screening.** In addition to the extractor and tagger, cancer screening involves the in-context classifier and attention aggregator modules. We applied the classifier (Eq. 4 in Supplementary Algorithm) to classify each test instance and applied the aggregator (Eq. 5 in Supplementary Algorithm) to integrate slide-level predictions. The recognition process was realized by similarity calculations in a non-parametric classification manner. The hyperparameter $k$ (default value 40) from Eq. 4 in Supplementary Algorithm was used to pick the top k similarities, which dynamically adapted each test slide according to the examples. Moreover, the hyperparameters within the aggregator encompassed the following variables: the temperature parameter $\tau$ for the softmax operation, $n$ denoting the number of top instance scores and $t_r$ indicating relevant instances. The default values were 10,2,000 and 0.88, respectively. When applied to diverse tasks, these values were grid-searched on the validation set. As shown in Extended Data Fig. 5b, these hyperparameters were robust, with the results fluctuating within 2.5%. Moreover, default values and those from other validation sets were also acceptable, as shown in Extended Data Fig. 5c.

**Cancer subtyping and tumor segmentation.** Compared to cancer screening, the subtyping task involves another instance miner module. As the normal issues are common across subtypes, we first need to locate the discriminative tumor regions. The miner is implemented by the basic tagger in Supplementary Algorithm. Different from the in-context miner for examples, we mine discriminative instances in the test slides; thus, the use of the uncertain range factor $t_e$ is different. We set it to 0.3 by default and searched for each task (Extended Data Fig. 5b). Other modules were the same as the screening task. As subtyping tasks have ≥2 cancer types, one-shot here indicates one example for each class. On the other hand, cancer screening and tumor segmentation tasks have one positive class; thus, only one cancer example was applied for one-shot. The example with the slide label prompt is an exception because, unlike other prompts, it cannot obtain negative samples outside the area outlined by the prompt; the slide label requires another normal slide as an example for one-shot. For the module design of the tumor segmentation task. There is another post-processor module based on the screening task. Specifically, it first reshapes the instance-level predictions to a prediction map and then conducts a Gaussian kernel to smooth the prediction map for a continuous tumor prediction. The bandwidth ($\sigma$) is set to 3 and the kernel side ($ks$) is 7, which is used as a sliding window to process the prediction map.

**Lymph node metastasis detection.** Metastasis detection is a special cancer screening task, where the metastatic tumors originate from other tissues and are very small and difficult to identify. The instance number ($n$) from Eq. 5 in Supplementary Algorithm is determined by the tumor size, which was set to 1 in CAMELYON datasets[28,30,31] to avoid involving non-tumor instances when the tumor is very tiny. Other hyperparameters were set to their default values. Furthermore, we used examples whose positive instances ranged from 1,000 to 3,000 to ensure example quality. When implementing the CAMLEYON17 (ref. [30]) and CAMELYON16-C (ref. [31]) datasets, we used examples from CAMELYON16 to directly assess the generalization capability of the benchmarks across various hospitals and scanning conditions. This evaluation protocol was also applied to other baselines for these two benchmarks. Further information on baseline method implementation is available in Supplementary Information, including few-shot, zero-shot and weakly supervised baselines.

## Pan-cancer dataset curation

**Dataset acquisition.** PRET is designed to perform pan-cancer recognition using a single model for multiple tasks and prompts without training. To demonstrate universality and wide applicability, we curated 23 benchmarks comprising a total of 4,484 WSIs containing 124 million local patches on an image size of 256 × 256 at ×20 magnification mainly for evaluation purposes, involving 18 cancer types and spanning nine major organs sourced from multiple international hospitals. These benchmarks covered diverse tasks, including cancer screening (ESCC, PTC, CRC, GC, LC, BC and PTC-QP), cancer subtyping (lymphoma, NSCLC-HQ, ESCA, sarcoma (SARC), RCC and NSCLC), tumor segmentation (ESCC, PTC, CRC, GC, LC, BC and PTC-QP) and lymph node metastasis detection (CAMELYON16, CAMELYON16-C and CAMELYON17). The main data source was GDPH, whereas QPCH represented the underserved region, where the GDP ranks 33rd of 34 first-level administration regions of China. Additionally, TCGA project and the CAMELYON series are public datasets from the USA and the Netherlands, respectively. The use of data from GDPH and QPCH was in compliance with institutional ethical standards and the Helsinki Declaration. The study protocols were approved by the respective institutional review boards of GDPH (approval no. KY2023-109-01) and QPCH (approval no. KYP-DL-202526), which waived informed consent for this retrospective study. Data from other sources were publicly available. The CAMELYON16 (ref. [28]) dataset was collected from two hospitals (RUMC and UMCU0), while CAMELYON17 (ref. [30]) involved three extra centers (RST, CWZ and LPON). The CAMELYON16-C (ref. [31]) dataset was built on CAMELYON16 to simulate scanning corruptions. The above hospitals are external data sources without pre-training or fine-tuning, except for TCAG pretrained by some foundation models. If one hospital provided examples for another hospital, there were also external and internal relations present (Fig. 5e). As shown in Fig. 1a, GDPH contributed 14 benchmarks: six cancer screening tasks, six tumor segmentation tasks and two cancer subtyping tasks. The QPCH provided two benchmarks related to PTC-QP for screening and segmentation, while hospitals from the Netherlands contributed three benchmarks for lymph node metastasis detection. Benchmarks from TCGA were collected from multiple sites from the USA, where four popular subtyping datasets are used: NSCLC, RCC, ESCA and SARC. Regarding randomization, case selection from GDPH and QPCH was retrospective and non-consecutive, targeting specific cancer types across defined periods (GDPH, 2020–2024; QPCH, 2024) to construct a representative pan-cancer benchmark. For all datasets—including GDPH, QPCH and public sources (TCGA, CAMELYON16, CAMELYON16-C and CAMELYON17)—case assignment to evaluate benchmarks was determined by intrinsic diagnostic and organ-type characteristics, without randomization. During few-shot evaluation, n-shot examples were randomly selected per run to ensure metric robustness. During the test, the researchers were blinded to the group allocation. No data points were excluded from the analyses after the datasets were finalized. For the internal datasets from GDPH and QPCH, WSIs that failed pre-defined quality thresholds (for example, because of excessive artifacts, poor staining, insufficient tissue or poor scanning quality) were excluded during initial benchmark construction.

Details of the dataset characteristics are provided in Extended Data Fig. 1 and Supplementary Table 1, including specific cancer types, basic demographics, tumor size statistics and WSI examples with visual prompts. To be specific, GDPH contributed 820 formalin-fixed paraffin-embedded (FFPE) H&E WSIs including both malignant and benign tissues. Benign tissue came from the esophagus (40), thyroid (41), colon (44), stomach (30), lung (47) and breast (41), while the cancer tissues included ESCC (60), PTC (61), CRC (59), GC (70), LC (58), BC (59), high-quality NSCLC (NSCLC-HQ; 97) and lymphoma (113). Note, the lymphoma dataset could be divided into two subtypes, diffuse large B cell lymphoma (64) and Hodgkin's lymphoma (49), while the NSCLC-HQ dataset contained lung adenocarcinoma (49) and lung squamous cell carcinoma (48). Additionally, QPCH provided another PTC dataset called PTC-QH, including 39 malignant slides and 47 benign WSIs. According to statistics in Extended Data Fig. 2a,b, the average age of participants from the proposed datasets was 54.4 years (8–90) and 50.6% of them were female on average. Leica Aperio-SS12181 scanners were used to scan data from these two centers at a pixel size of 0.2642 $\mu$m. Only de-identified retrospective data were used for research after we eliminated sensitive participant data, including name, birthdate, admission year and admission number.

Public datasets included 2,488 H&E FFPE WSIs from TCGA project, encompassing ESCA (149), SARC (346), RCC (939) and NSCLC (1,052). ESCA included ESCC (86) and esophageal adenocarcinoma (65), while SARC included leiomyosarcoma, not otherwise specified (152), dedifferentiated liposarcoma (87) and undifferentiated sarcoma (107). RCC subtypes included kidney chromophobe (121), kidney renal clear cell carcinoma (519) and kidney renal papillary cell carcinoma (299). NSCLC subtypes included lung adenocarcinoma (540) and lung squamous cell carcinoma (512). For the lymph node metastasis detection of BC, there were 399 WSIs in CAMELYON16 (129 WSIs as the standard test set), which were augmented to simulate scanning corruption in CAMELYON16-C (399), while 499 WSIs from the CAMELYON17 dataset were used for evaluation from five Netherlands hospitals.

**Visual prompt labeling.** To enhance the flexibility and efficiency of annotation, we introduced multiple types of visual prompts: L, B, R and M. These diverse labeling approaches balance pathologists' annotation budgets and optimize task performance. Slide labels, commonly used in weakly supervised methods such as multiple instance learning[10,18,37], provide slide-level class label. Tumor masks offer instance-level annotations using labeled points on precise tumor boundaries to outline tumor regions. Bounding boxes enclose tumor regions with minimal non-tumor areas, where only two points are required on the left top corner and right bottom corner, respectively. The rough mask is a label-efficient prompt, where fewer points are labeled compared to the mask label. Moreover, the points are not required to be precise on the tumor boundaries; instead, they are labeled outside tumors and roughly enclose them. Note that our visual prompts do not rely on an extra model[54] or clustering step [55] to label the regions of interest. Representative visual prompts are shown in Extended Data Fig. 2d.

We obtained the image label from diagnosis records without manual effort, except for partial cases to account for labeling time of new slides. Other prompt types were labeled by 11 certified pathologists (Z.N., Z.L., F.F., B.Z., X.Q., L.S., J.Q., L.X., C.F., T.Q. and Q.W.) independently following the criteria introduced above. We provided these visual prompts for datasets from GDPH, PQCH and TCGA, whereas CAMELYON16 (ref. [28]) released its tumor masks. For the cancer screening task and subtyping task, we labeled around 20 WSIs with four visual prompt types for each cancer type. These labeled slides were used as examples at quantities of 1–8, where 20 prompts were sufficient to ensure example diversity. These prompts were released. On the one hand, they can be directly used for deployment in other hospitals. As shown in Fig. 5e, these prompts showed good generalization across external hospitals. On the other hand, they can be used as ground truth to validate the interpretability and segmentation effectiveness of existing methods.

Other labeling details are shown in Fig. 1c. To be specific, the average labeling time was 15.4 s. For the tumor mask, pathologists precisely drew the boundaries of the tumor, excluding normal tissues, which took the most time, and the average labeling time is 537.2 s. Instead, pathologists used crude drawings or boxes to identify the tumor location in the other two labels, which might have included benign tissues. With corresponding average labeling times of 85.1 and 44.6 s for the rough mask and bounding box, respectively, these two labels were more efficient in labeling costs compared to the tumor mask and showed small performance gaps compared to the tumor masks in the segmentation task as shown in Fig. 3d. We also counted the tumor size as a density map in Extended Data Fig. 2c, where most fractions of PTC, PTC-QP, CAMELYON16 and ESCC were located to the left of the x axis, indicating that most tumors were small with a small number of patches. We found that those datasets with small tumors needed stronger visual prompts for better performances, allowing PRET to excel as rich local information was fully used.

**Data split.** The dataset split mainly included three cases, as shown in Supplementary Table 1. For TCGA benchmarks, the common split protocol was to use 70% of the data for training, 10% for validation and 20% for testing. As PRET does not involve training, we randomly selected 20% of WSIs from the whole dataset using a fixed list and random seed (1,024) as the test set for each independent experiment, with validation sets of 100 WSIs, or 50 cases when data were insufficient (ESCA). Example WSIs were randomly selected from annotated data with visual prompts, with 1–8 examples chosen for inference in context. When sampling examples, we made sure that each example set was different from other independent experiments to ensure diverse and stable example quality. Then, we divided the remaining slides into validation and test sets at target numbers, ensuring that the participants in the example slides did not overlap with the test set to prevent data leakage. Furthermore, the proposed benchmarks were specifically designed for pan-cancer recognition without many training samples in conventional datasets. We allocated two thirds of the remaining WSIs as the test set and one third as the validation set after random example selection to fully use the examples and ensure a low P value for each repeat experiment. CAMELYON benchmarks had a fixed dataset split, with CAMELYON16 and CAMELYON16-C having a fixed test set of n = 129. For the training set, examples were randomly selected from WSIs with prompts and 100 examples were randomly sampled from slides without prompts. In CAMELYON17, all 499 WSIs were considered the test set, while the validation set and examples were from CAMELYON16. This split simulated deploying CAMELYON16 models directly to CAMELYON17 and CAMELYON16-C to verify the model's generalization across hospital examples. All these data splits are reproducible as the datasets and data lists were released under a fixed random seed. For all implemented baselines, the data splits mentioned above were the same to ensure fair comparisons.

### Computing software and hardware

We used Python (version 3.8.0) and PyTorch (version 1.12.1; https://pytorch.org) to build this project. We use the ImageScope to annotate the visual prompt (www.leicabiosystems.com/digital-pathology/manage/aperio-imagescope). The WSI was sliced into patches using viplibs (www.libvips.org). We read the WSI using openslide (version 1.4.1; https://openslide.org/) and visualized the data using matplotlib (version 3.5.0; https://matplotlib.org/). We used opencv-python (version 4.4.0; https://pypi.org/project/opencv-python/) in the tagger and miner for the OTSU binarization and processing of patch images. For evaluation, we used metrics through torchmetrics (version 0.9.1; https://lightning.ai/docs/torchmetrics/) and scikit-learn (version 0.22.1). The Wilcoxon test was implemented using scipy (version 1.7.3; https://scipy.org/). For the feature encoder, the ViT [48] architectures were implemented through timm library (version 0.9.6; https://github.com/huggingface/pytorch-image-models). The codebase of MIL baselines was built using CLAM (https://github.com/mahmoodlab/CLAM), integrating other baselines from their official codes. Our method only requires relatively few computing resources, including an NVIDIA 3090 GPU to extract features and compute cosine similarities. The CPU used was an Intel(R) Xeon(R) Platinum 8375C CPU @ 2.90 GHz.

### Evaluation metrics

We mainly report the results evaluated by the AUC metric, specifically derived from the receiver operating characteristic (ROC) curve. For binary tasks, the AUC was computed by plotting the true positive rate (sensitivity) against the false positive rate (1 − specificity) with varied classification thresholds. Extending to multiclass tasks, the AUC indicates the macro average of AUC values of each class, which is computed by taking the unweighted average values across classes. Furthermore, we report the accuracy and F1 score in Supplementary Tables 2–9 for comprehensive evaluations. Similarly, we applied the macro accuracy and macro F1 score for multiclass tasks (also referred to as the balanced accuracy and balanced F1 score, respectively). For the segmentation task, we used the Dice metric to evaluate the degree of overlap between the prediction mask A and ground truth B as $\mathrm{Dice}(A,B)=\frac{2\times|A\cap B|}{|A|+|B|}$, which is equal to the F1 score for instances in the binary case. The accuracy, F1 score and Dice thresholds were determined from the validation set on the basis of the highest accuracy or Dice value achieved. This protocol was applied across all baselines to ensure fair comparisons.

**Statistics and reproducibility**

No statistical methods were used to pre-determine sample sizes but our sample sizes are similar to those reported in previous publications[11,16,41]. Data distribution was assumed to be normal but this was not formally tested. Non-parametric tests (two-sided Wilcoxon test) were used for all analyses. For each result from PRET and implemented baselines, we conducted independent experiments five times. When using four different prompts, we increased the number of experiments to n = 20. Specifically, when evaluating the performance of each prompt, we ran five times the benchmark number of experiments. The reported values are the mean ± s.d. of these repeated experiments. The observed differences in model performance compared to the best-performing baseline were tested for statistical significance using a two-sided Wilcoxon test. The P value is the proportion of differences in model performance greater than the observed difference in terms of absolute value. For multiple individual experiments, we report the median of involved P values. The null hypothesis was that there is no difference in model performance for the given test set and evaluation metric.

Split protocols were as described above. During the test, the researchers were blinded to the group allocation.

**Data availability**

The in-house datasets from GDPH and QPCH are publicly available (https://huggingface.co/datasets/yili7eli/PRET/tree/main), including ESCC, PTC, CRC, GC, LC, BC, lymphoma, NSCLC-HQ and PTC-QP. The visual prompts involved for both in-house datasets and open datasets were also released, with data lists to reproduce data splits. The public datasets CAMELYON16 (ref. [28]) and CAMELYON17 (ref. [30]) are available online (https://camelyon16.grand-challenge.org/) and CAMELYON16-C (ref. [31]) was realized by scanning corruption with code available from GitHub (https://github.com/superjamessyx/robustness_benchmark). TCGA datasets can be found from the National Institutes of Health Genomic Data Commons (https://portal.gdc.cancer.gov/), including NSCLC, RCC, ESCA and SARC. Source data are provided with this paper.

**Code availability**

All the involved model weights and Python packages are available online. Our work is publicly available from GitHub (https://github.com/xmed-lab/PRET), with detailed instructions, comments and evaluation scripts.

## Acknowledgements

Y.L., T.X., Qixiang Zhang and X.L. received support from the Research Grants Council (RGC) of the Hong Kong Special Administrative Region (project nos. R6005-24, AoE/E-601/24-N and T45-401/22-N), the Hong Kong Joint Research Scheme of the National Natural Science Foundation of China/RGC (project no. N_HKUST654/24), the Hong Kong Innovation and Technology Fund under Project PRP/041/22FX and the National Natural Science Foundation of China (grant no. 62306254). T.X. also received support from the Hong Kong PhD Fellowship Scheme. Z.N. and Qingling Zhang are supported by grants from the Key R&D Program Projects in Guangdong Province (2021B0101420005 to Qingling Zhang), the National Natural Science Foundation of China (grant no. 82173033 to Qingling Zhang), the High-level Hospital Construction Project (DFJHBF202108 and YKY-KF202204 to X.-W.B. and Qingling Zhang) and the Guangdong Provincial Key Laboratory of AI in Medical Image Analysis and Application (2022B1212010011 to Z.L.). K.-H.Y. is supported in part by the National Institute of General Medical Sciences (grant no. R35GM142879), the National Heart, Lung and Blood Institute (grant no. R01HL174679), the Department of Defense Peer-Reviewed Cancer Research Program Career Development Award (HT9425-231-0523), the Research Scholar Grant (RSG-24-1253761-01-ESED) to K.-H.Y. from the American Cancer Society and the Harvard Medical School Dean's Innovation Award. K.Z. is supported by the National Natural Science Foundation of China (W2431057), Macau Science and Technology Development Fund, Macau (0007/2020/AFJ, 0070/2020/A2 and 0003/2021/AKP) and Guangzhou National Laboratory (YW-SLJC0201). We would like to thank Q. Shao from the Hong Kong University of Science and Technology for his valuable suggestions.

## Author contributions

X.L. and Qingling Zhang planned the study. Y.L. designed the PRET methods guided by X.L. and conducted the experiments. Z.N. and Qingling Zhang collected the in-house data required for this study and organized the data labeling. T.X. and Qixiang Zhang implemented some baseline methods. Y.L., X.L., Qingling Zhang and K.-H.Y contributed to experimental design. The in-house WSIs were collected by Z.N. and Qingling Zhang, with support from M.Y. and X.-W.B. Data labeling was performed by Z.N., Z.L., F.F., B.Z., X.Q., L.S., J.Q., L.X., C.F., T.Q. and Q.W., who also accounted for the labeling costs. X.L. supervised the study. Y.L. wrote the paper, with contributions from Z.N., X.L. and K.Z. All authors discussed the results and contributed to the final manuscript.

## Competing Interests

Qingling Zhang, Z.N., X.L. and Y.L. are inventors on a pending Chinese patent application (application no. 202510130037.4, 'A pan-cancer AI pathology diagnosis system using few examples without training'; applicant: GDPH, Guangdong Academy of Medical Sciences) related to the PRET algorithm described in this paper. K.-H.Y. is an inventor on US patent 10,832,406. This patent is assigned to Harvard University and is not directly related to this paper. The other authors declare no competing interests.

## References

1. Ferlay, J. *et al.* Cancer statistics for the year 2020: an overview. *Int.* J. C*ancer* **149**,778–789 (2021).
2. Benediktsson, H., Whitelaw, J. & Roy, I. Pathology services in developing countries: a challenge. *Arch.* P*athol*. L*ab*. M*ed*. **131**,1636–1639 (2007).
3. Metter, D. M., Colgan, T. J., Leung, S. T., Timmons, C. F. & Park, J. Y. Trends in the US and Canadian pathologist workforces from 2007 to 2017. *JAMA* N*etw*. O*pen* **2**, e194337 (2019).
4. Märkl, B., Füzesi, L., Huss, R., Bauer, S. & Schaller, T. Number of pathologists in Germany: comparison with European countries, USA, and Canada. *Virchows Arch.* **478**,335–341 (2021).
5. Bera, K., Schalper, K. A., Rimm, D. L., Velcheti, V. & Madabhushi, A. Artificial intelligence in digital pathology—new tools for diagnosis and precision oncology. *Nat.* R*ev*. *Clin.* O*ncol*. **16**,703–715 (2019).
6. Song, A. H. *et al.* Artificial intelligence for digital and computational pathology. *Nat. Rev. Bioeng.* **1**,930–949 (2023).
7. Campanella, G. *et al.* Clinical-grade computational pathology using weakly supervised deep learning on whole slide images. *Nat.* M*ed*. **25**,1301–1309 (2019).
8. Huang, S.-C. *et al.* Deep neural network trained on gigapixel images improves lymph node metastasis detection in clinical settings. *Nat.* C*ommun*. **13**,3347 (2022).
9. Kundra, R. *et al.* OncoTree: a cancer classification system for precision oncology. *JCO* C*lin*. C*ancer* I*nform*. **5**,221–230 (2021).
10. Lu, M. Y. *et al.* Data-efficient and weakly supervised computational pathology on whole-slide images. *Nat.* B*iomed*. E*ng*. **5**,555–570 (2021).
11. Lu, M. Y. *et al.* A visual-language foundation model for computational pathology. *Nat. Med*. **30**,863–874 (2024).
12. Wang, X. *et al.* A pathology foundation model for cancer diagnosis and prognosis prediction. *Nature* 634,970–978 (2024).
13. Xu, H. *et al.* A whole-slide foundation model for digital pathology from real-world data. *Nature* **630**,181–188 (2024).
14. Arslan, S. *et al.* A systematic pan-cancer study on deep learning-based prediction of multi-omic biomarkers from routine pathology images. *Commun. Med*. **4**,48 (2024).
15. Huang, Z., Bianchi, F., Yuksekgonul, M., Montine, T. J. & Zou, J. A visual–language foundation model for pathology image analysis using medical Twitter. *Nat.* M*ed*. **29**,2307–2316 (2023).
16. Chen, R. J. *et al.* Towards a general-purpose foundation model for computational pathology. *Nat. Med*. **30**,850–862 (2024).
17. Kang, M., Song, H., Park, S., Yoo, D. & Pereira, S. Benchmarking self-supervised learning on diverse pathology datasets. In *Proc. IEEE/CVF Conference on Computer Vision and Pattern Recognition* (eds Brown, M. S. et al.) 3344–3354 (IEEE, 2023).
18. Ilse, M., Tomczak, J. & Welling, M. Attention-based deep multiple instance learning. In Proc. 35th *International* C*onference on* M*achine* L*earning* (eds Dy, J. & Krause, A.) 2127–2136 (PMLR, 2018).

19. Chen, R. J. *et al.* Scaling vision transformers to gigapixel images via hierarchical self-supervised learning. In *Proc. IEEE/CVF Conference on Computer Vision and Pattern Recognition* (eds Chellappa, R. et al.) 16144–16155 (IEEE, 2022).

20. Snell, J., Swersky, K. & Zemel, R. Prototypical networks for few-shot learning. In Proc. 31st I*n*ternational Conference on N*eural* I*nformation* P*rocessing* S*ystems* (eds von Luxburg, U. et al.) 4080–409**0** (ACM, 2017).

21. Wang, Y., Chao, W.-L., Weinberger, K. Q. & van der Maaten, L. SimpleShot: revisiting nearest-neighbor classification for few-shot learning. P*reprint a*t https://arxiv.org/abs/*1911.04623* (2019).

22. Brown, T. *et al.* Language models are few-shot learners. In Proc. 34th I*n*ternational Conference on N*eural* I*nformation* P*rocessing* S*ystems* (eds Larochelle, H. et al.) 1877–1901 (Curran Associates, Inc., 2020).

23. Wang, X., Wang, W., Cao, Y., Shen, C. & Huang, T. Images speak in images: a generalist painter for in-context visual learning. In *Proc. IEEE/CVF Conference on Computer Vision and Pattern Recognition* (eds Brown, M. S. et al.) 6830–6839 (IEEE, 2023).

24. Zhang, J., Wang, B., Li, L., Nakashima, Y. & Nagahara, H. Instruct me more! Random prompting for visual in-context learning. In *Proc. IEEE/CVF Winter Conference on Applications of Computer Vision* (eds Souvenir, R. et al.) 2585–2594 (IEEE, 2024).

25. Sheng, D. *et al.* Towards more unified in-context visual understanding. In *Proc. IEEE/CVF Conference on Computer Vision and Pattern Recognition* (eds Camps, O. et al.) 13362–13372 (IEEE, 2024).

26. Zhao, H. *et al.* MMICL: empowering vision-language model with multi-modal in-context learning. In Proc. 12*th International Conference on Learning Representations* (ed. Kim, B.) (ICLR, 2024).

27. Ferber, D. *et al.* In-context learning enables multi-modal large language models to classify cancer pathology images. *Nat. Commun.* **15**,10104 (2024).

28. Bejnordi, B. E. *et al.* Diagnostic assessment of deep learning algorithms for detection of lymph node metastases in women with breast cancer. *J*AMA **318**,2199–2210 (2017).

29. Qu, L. *et al.* The rise of AI language pathologists: exploring two-level prompt learning for few-shot weakly-supervised whole slide image classification. In Proc. 37th International Conference on *Neural Inf*ormation *Process*ing *Syst*ems (eds Oh, A. et al.) **6**7551–67564 (Curran Associates, Inc., 2023).

30. Litjens, G. *et al.* 1399 H&E-stained sentinel lymph node sections of breast cancer patients: the CAMELYON dataset. *Gigascience* **7**, giy065 (2018).

31. Zhang, Y. *et al.* Benchmarking the robustness of deep neural networks to common corruptions in digital pathology. In Proc. 25th *International Conference on Medical Image Computing and Computer-Assisted Intervention* (eds Wang, L. et al.) 242–252 (Springer, 2022).

32. He, K., Zhang, X., Ren, S. & Sun, J. Deep residual learning for image recognition. In *Proc. IEEE* C*onference on* C*omputer* V*ision and* P*attern* R*ecognition* (eds Tuytelaars, T. et al.) 770–778 (IEEE, 2016).

33. Deng, J. *et al.* ImageNet: a large-scale hierarchical image database. In Proc. *IEEE* C*onference on* C*omputer* V*ision and* P*attern* R*ecognition* (eds Huttenlocher, D. et al.) 248–255 (IEEE, 2009).

34. Vorontsov, E. *et al.* A foundation model for clinical-grade computational pathology and rare cancers detection. *Nat.* M*ed*. **30**,2924–2935 (2024).

35. Ma, J. *et al.* A generalizable pathology foundation model using a unified knowledge distillation pre-training framework. *Nat. Biomed. Eng.* https://doi.org/10.1038/s41551-025-01488-4 (2025).

36. Ding, T. *et al.* A multi-modal whole-slide foundation model for pathology. *Nat*. M*ed*. 31,3749–3761 (2025).

37. Shao, Z. *et al.* TransMIL: transformer based correlated multiple instance learning for whole slide image classification. In Proc*.* 35th I*n*ternational Conference on N*eural* I*nformation* P*rocessing* S*ystems* (eds Ranzato, M. et al.) 2136–2147 (Curran Associates, Inc., 2021).

38. Li, H. *et al.* Task-specific fine-tuning via variational information bottleneck for weakly-supervised pathology whole slide image classification. In *Proc. IEEE/CVF Conference on Computer Vision and Pattern Recognition* (eds Brown, M. S. et al.) 7454–7463 (IEEE, 2023).

39. Tang, W. *et al.* Multiple instance learning framework with masked hard instance mining for whole slide image classification. In *Proc. IEEE/CVF Conference on Computer Vision* and Pattern Recognition (eds Brown, M. S. et al.) 4078–4087 (IEEE, 2023).

40. Chen, Y. *et al.* dMIL-Transformer: multiple instance learning via integrating morphological and spatial information for lymph node metastasis classification. *IEEE J. Biomed. Heal*th *Inform*. **27**,4433–4443 (2023).

41. Xiang, J. *et al.* A vision-language foundation model for precision oncology. *Nature* 638,769–778 (2025).

42. Coudray, N. *et al.* Classification and mutation prediction from non-small cell lung cancer histopathology images using deep learning. *Nat.* M*ed*. **24**,1559–1567 (2018).

43. Liang, J. *et al.* Deep learning supported discovery of biomarkers for clinical prognosis of liver cancer. *Nat. Mach. Intell.* **5**,408–420 (2023).

44. Zhou, Y., Li, X., Wang, Q. & Shen, J. Visual in-context learning for large vision-language models. In *Findings of the Association for Computational Linguistics* (eds Ku, L.-W. et al.) 15890–15902 (*ACL*, 2024).

45. Li, C. *et al.* LLaVA-med: training a large language-and-vision assistant for biomedicine in one day. In Proc*.* 37th International Conference on *Neural Inf*ormation *Process*ing *Syst*ems (eds Oh, A. et al.) 28541–285**6**4 (Curran Associates, Inc., 2023).

46. Li, Y. *et al.* Few-shot lymph node metastasis classification meets high performance on whole slide images via the informative non-parametric classifier. In Proc. 25th *International Conference on Medical Image Computing and Computer-Assisted Intervention* (eds Linguraru, M. G. et al.) 109–119 (Springer, 2024).

47. Caron, M. *et al.* Emerging properties in self-supervised vision transformers. In *Proc*. *IEEE/CVF* I*nternational* C*onference on* C*omputer* V*ision* (eds Berg, T. et al.) 9650–9660 (IEEE, 2021).

48. Dosovitskiy, A. *et al.* An image is worth 16x16 words: Transformers for image recognition at scale. In Proc. 9th *International Conference on Learning Representations* (ed. Mohamed, S.) (ICLR, 2021).

49. Xu, Y. *et al.* A multi-modal knowledge-enhanced whole-slide pathology foundation model. *Nat*. *Commun*. 16,11406 (2025).

50. Wang, X. *et al.* Transformer-based unsupervised contrastive learning for histopathological image classification. *Med.* I*mage* A*nal*. **81**,102559 (2022).

51. Oquab, M. *et al.* DINOv2: Learning robust visual features without supervision. *Trans*. *Mach. Learn. Res.* https://openreview*.* net/forum?id=a68SUt6zFt (2024).

52. Radford, A. *et al.* Learning transferable visual models from natural language supervision. In Proc. *International* Conference *on* Machine *Learning* (eds Meila, M. & Zhang, T.) 8748–8763 (PMLR, 2021).

53. Ding, J. *et al.* LongNet: scaling transformers to 1,000,000,000 tokens. *In Proc. 10th International Conference on Learning Representations* (eds Hofmann, K. & Rush, A.) (ICLR, 2023).

54. Komura, D. *et al.* Universal encoding of pan-cancer histology by deep texture representations. *Cell Rep*. **38**,110424 (2022).

55. Riasatian, A. *et al.* Fine-tuning and training of DenseNet for histopathology image representation using tcga diagnostic slides. *Med.* I*mage Anal*. **70**,102032 (2021).

56. Lu, M. Y. *et al.* AI-based pathology predicts origins for cancers of unknown primary. *Nature* **594**,106–110 (2021).

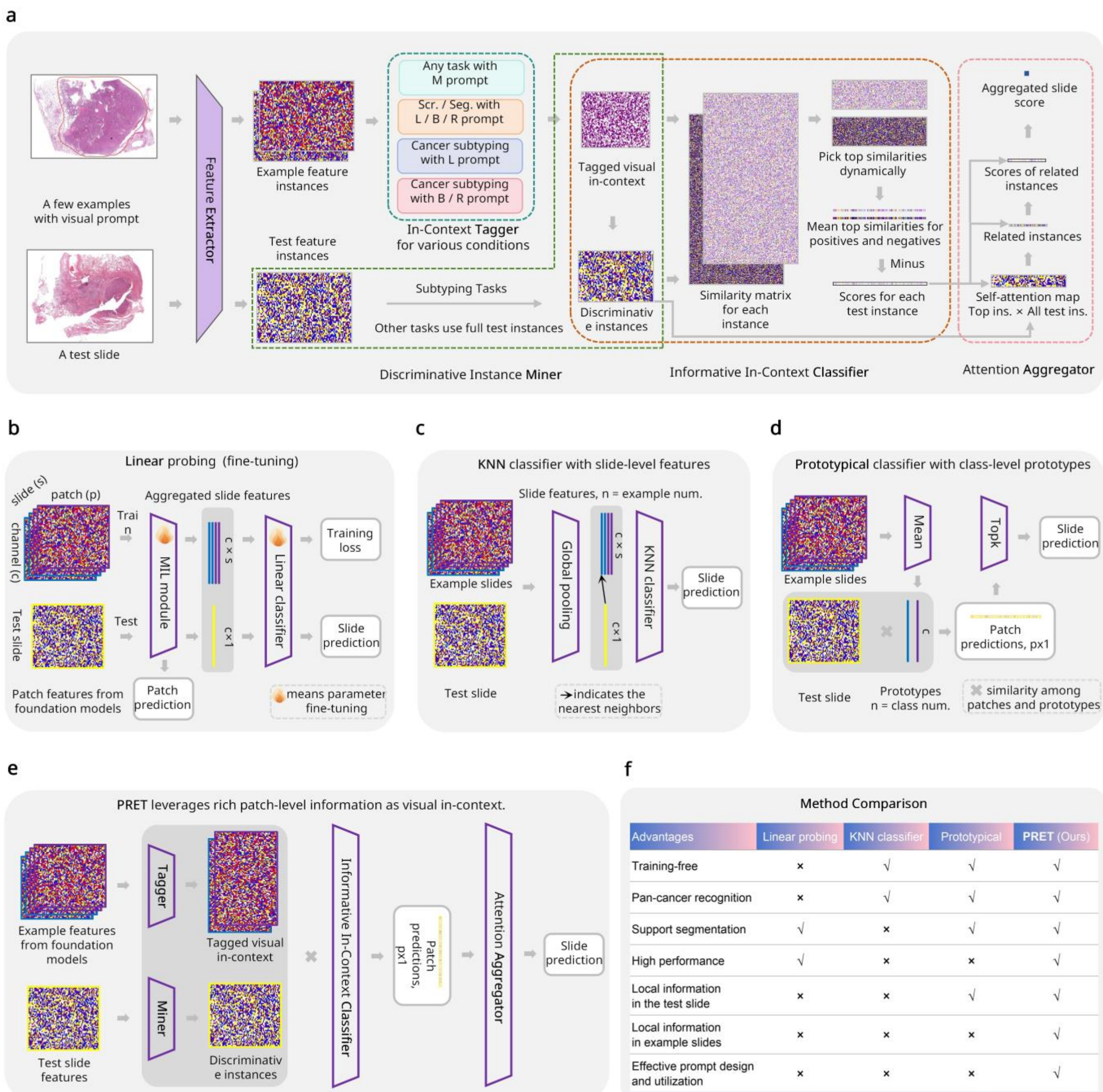

Extended Data Fig. 1 | Method details of PRET and comparison with other settings. a. PRET contains six major components, including the extractor, tagger, miner, classifier, aggregator, and post processor. b. Linear probing methods require task-specific parameter fine-tuning. The fine-tuned parameters exist in the multiple instance learning (MIL) module and linear classifier. Both the training and testing examples are embedded into slide-level global features. c. The KNN classifier also deploys slide-level features for each example slide without local path-level features. The MIL module turns to be global pooling from and linear classifier becomes KNN classifier. d. The MI-Prototype20 and MI-SimpleShot16 apply the prototypical classifier, where the prototypes are the mean features from examples. Note that prototypes are dataset-level features whose size is the number of classes. Then, they are used to calculate similarities with patch-level features from the test slide with a top K operation. e. The proposed PRET preserves rich information of patch-level features from both example and test slides, with novel modules to leverage visual in-context. The new modules involve a visual in-context tagger to process visual prompts, a miner to explore discriminative test patches, an informative in-context classifier to classify patches, and an attention aggregator to get the slide-level prediction. f. Comparison among methods in multiple aspects. PRET is unique at local information in example slides, effective prompt design and utilization, with advantages in aspects of training-free, pan-cancer recognition, segmentation supporting, high performance, local information in the test slide.

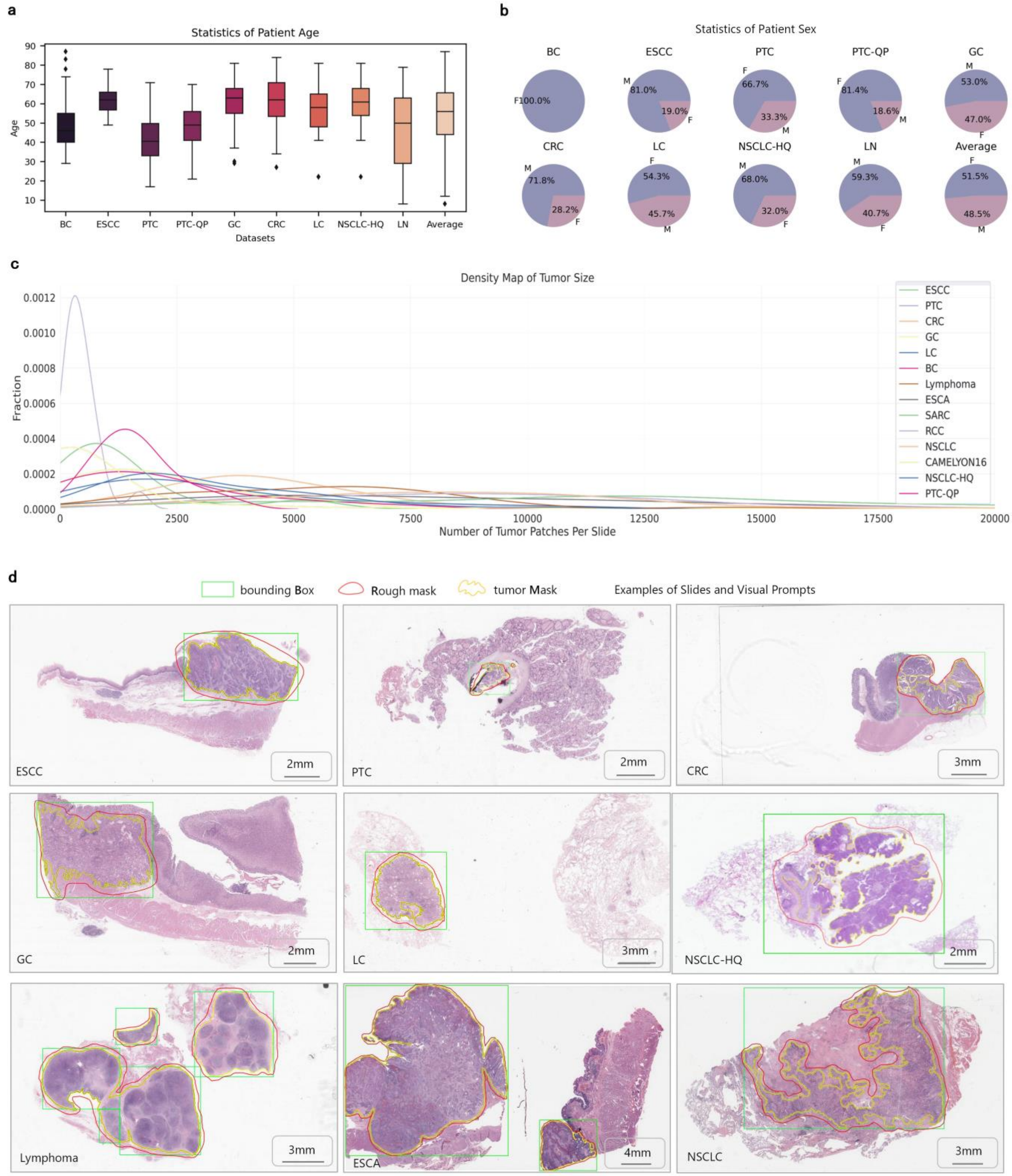

Extended Data Fig. 2 | Dataset characteristics and labeling details. a. Statistics of patient age for the proposed datasets (n = 906). b. Statistics of age patient sex for the proposed datasets (n = 906). c. Density map of tumor size (n = 565). The x-axis indicates the number of tumor patches per slide, suggesting the tumor size with the frequency on the y-axis. d. Examples of slides and visual prompts. The slide label is a certain number, while other visual prompts are boxes or outlines given in these examples.

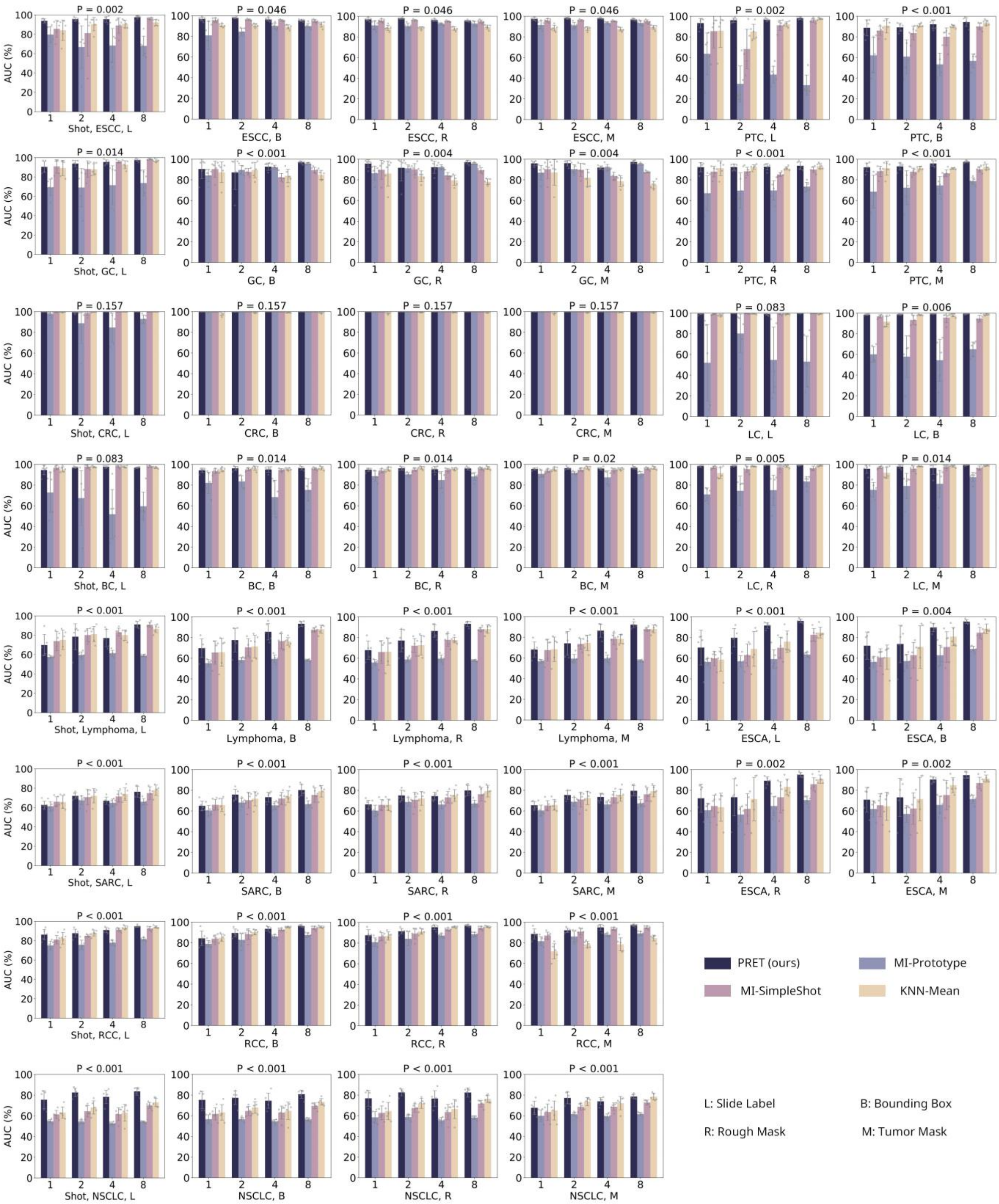

Extended Data Fig. 3 | Pan-cancer recognition performances with multiple visual prompts. The proposed PRET almost surpasses the baselines in all datasets and prompts at the AUC metric. The gray error bars indicate standard deviation, and the gray dots are the specific results for n = 5 independent experiments. The p value indicates the significance of PRET outperforming the best baseline with two-sided Wilcoxon test, reporting the median value across all repeat experiments and varied shots.

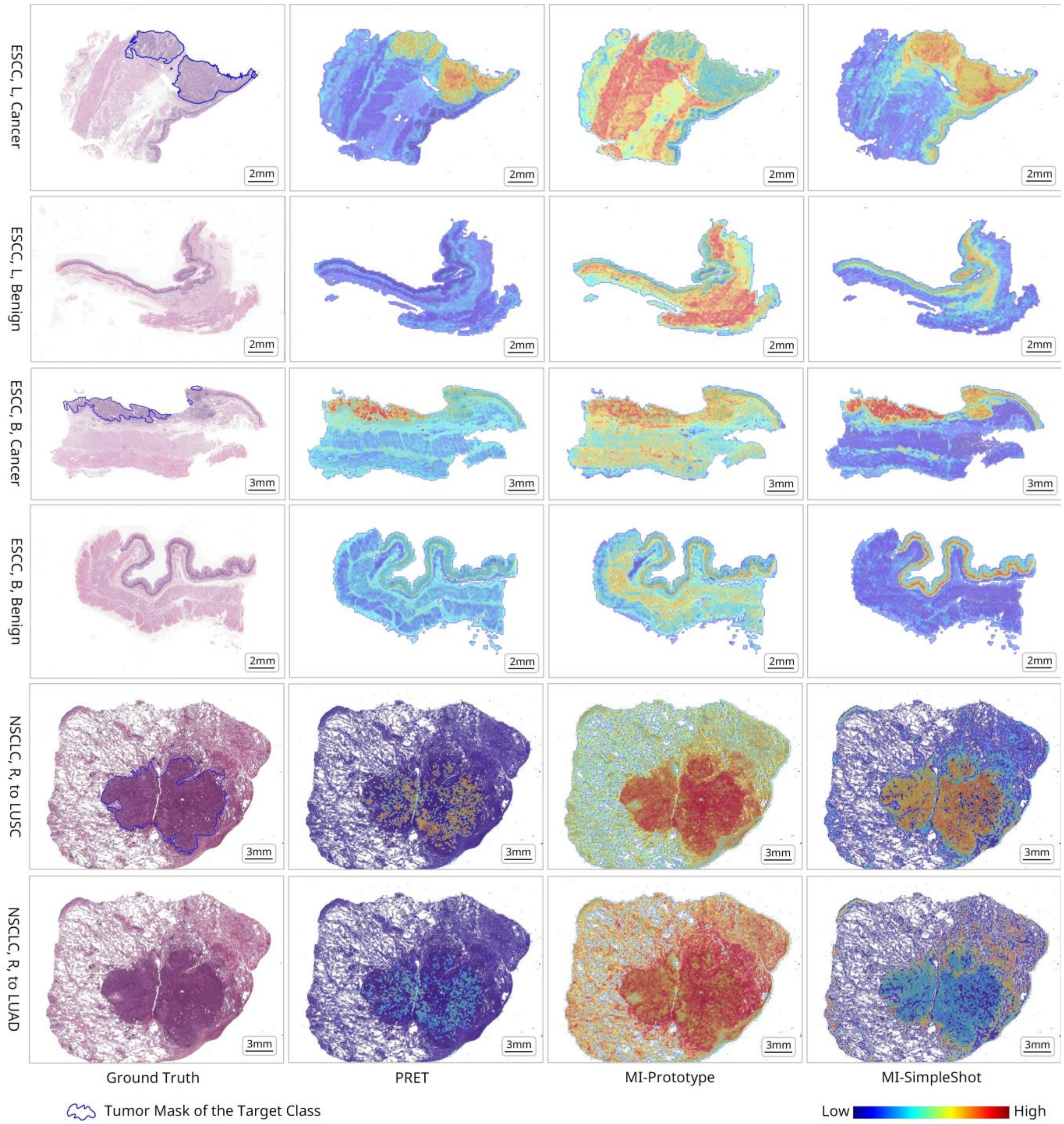

Extended Data Fig. 4 | Visualization and case study for cancer screening and subtyping. The classification tasks do not require accurate pixel-level localization as the segmentation task. Instead, the score maps are supposed to be distinguishable among classes. The score maps between positive and negative examples are much more distinguishable for the proposed PRET compared with other methods. So that PRET meets fewer false positives. These cases include ESCC for cancer screening and NSCLC for cancer subtyping, involving multiple visual prompt types, where L indicates slide label and B, R means bounding box and rough mask, respectively.

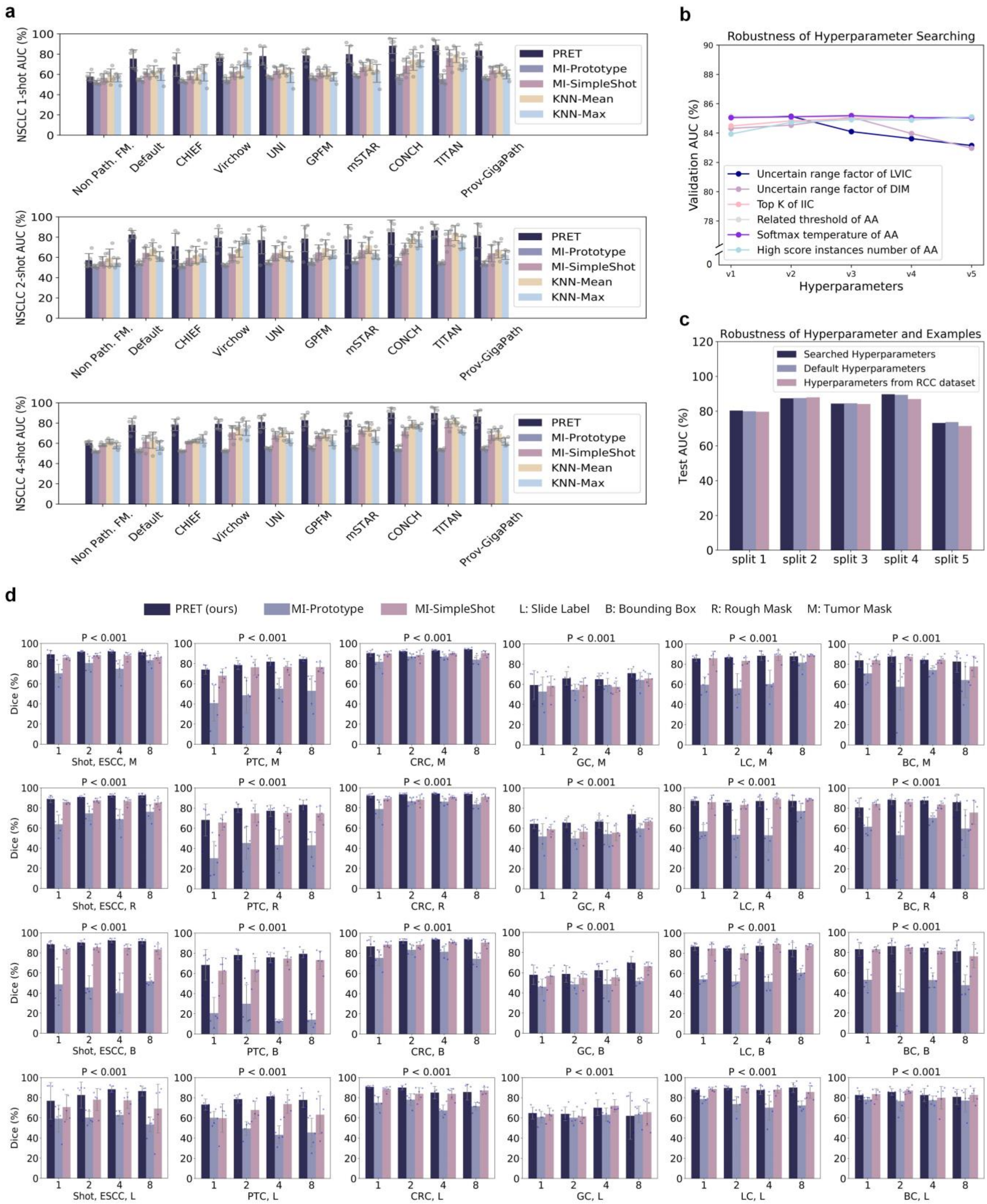

Extended Data Fig. 5 | Experiment results about foundation models, hyperparameters, and segmentation. a. PRET is born for foundation models with much larger performance gaps compared with the non-foundation model (ResNet-5032 pretrained from ImageNet). There are 1-4 shot results. The error bars show the standard deviation across n = 5 independent experiments, and the gray dots are the specific results for different data splits. b. Hyperparameter gird search on NSCLC validation set with slide labels at 8-shot. The result fluctuates within 2.5%, showcasing the model's robustness. Among

these hyperparameters, the uncertain range factor te in Supplementary Algorithm 1 for the local visual in-context tagger (LVIT) and discriminative instance miner (DIM) is relatively influential since they control the quantity of visual in-context. {v1, v2, v3, v4, v5} are {0,0.02,0.04,0.06,0.08} for the uncertain range factor of the LVIT in the examples, and {0.1,0.15,0.2,0.25,0.3} for that of the DIM to the test slides, {20,30,40,50,60} for top k of the informative in-context classifier (IIC). Besides, they are set to {0.86,0.87,0.88,0.89,0.9}, {1000,2000,3000,4000,5000}, {1,5, 10,20,30} for the related threshold tr, high score instances number n, and softmax temperature τ in the attention aggregator (AA). c. The results on different data splits and hyperparameter groups. Thanks to the robustness, default hyperparameters without grid search and those from the RCC dataset also perform well, showing close performances compared with the searched hyperparameters group. Different data splits reflect the varied example quality, which is less robust than hyperparameters, thus we report the average results for better stability. d. The proposed PRET almost surpasses the baselines in all datasets and prompts at the DICE metric. The gray error bars indicate standard deviation across n = 5 independent experiments, and the gray dots are the specific results for different data splits. The p value indicates the significance of PRET outperforming the best baseline with two-sided Wilcoxon test, reporting the median value across all repeat experiments and varied shots.

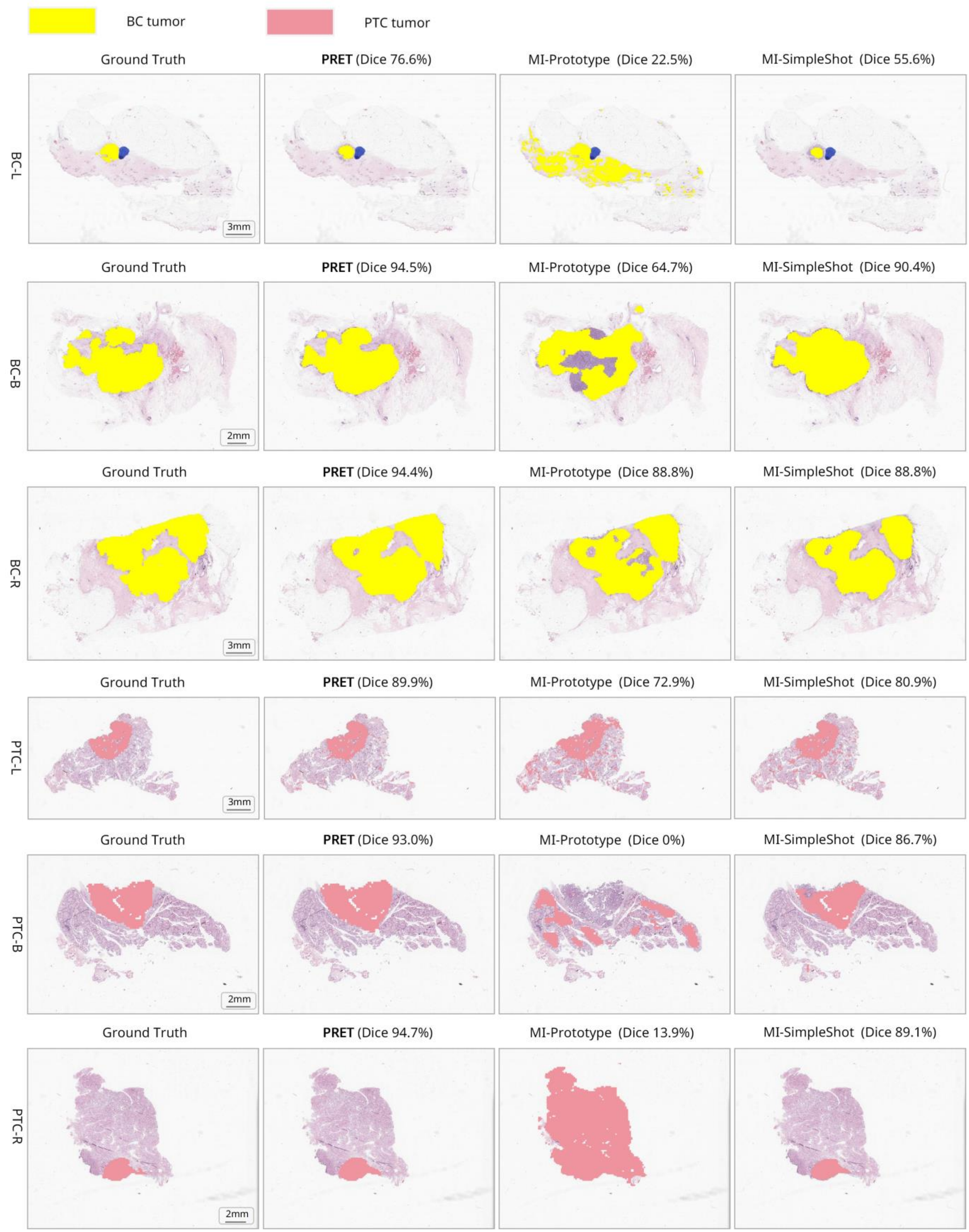

Extended Data Fig. 6 | Visualization and comparison of tumor segmentation task based on weak visual prompts. Qualitative illustrations comparing PRET to MI-Prototype and MI-SimpleShot on PTC and BC datasets using 8 examples. The tumor region is colored yellow. Note that the blue area in the first "BC-L" row are stains on the slide from a blue marker pen.

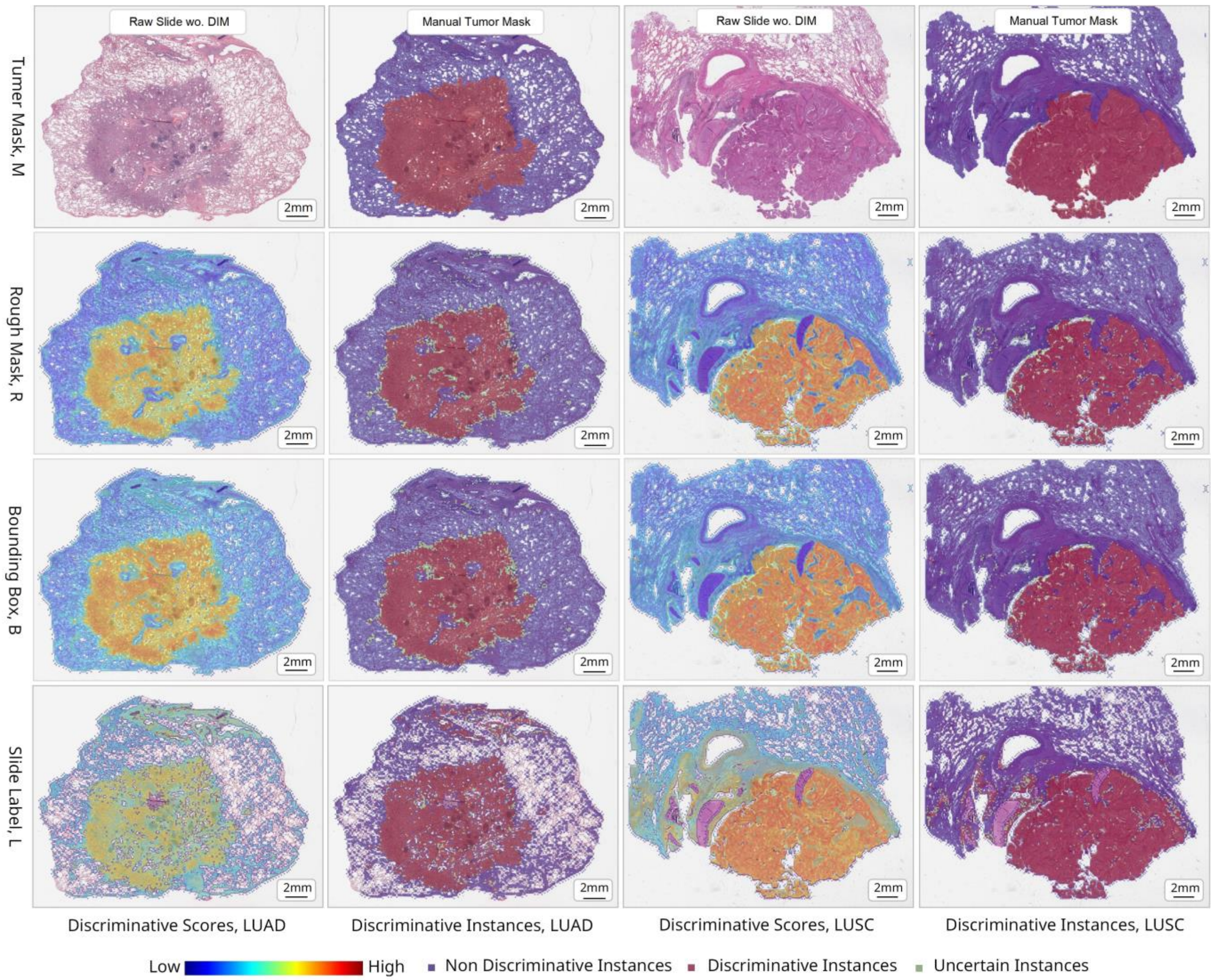

Extended Data Fig. 7 | Visualization of in-context tagger. Our method successfully tags the instances of examples that are close to manually labeled regions via weak visual prompts. This in-context tagger module is used for examples to label visual in-context instead of weakly segmentation for the test slides. Different visual prompts support varied conditions and provide more choices for pathologists. The bounding box and rough mask show fewer responses out of the manual mask with better score maps, while the slide label produces more positive regions. The used dataset is NSCLC (subtyping for LUAD and LUSC) on 8-shot settings.

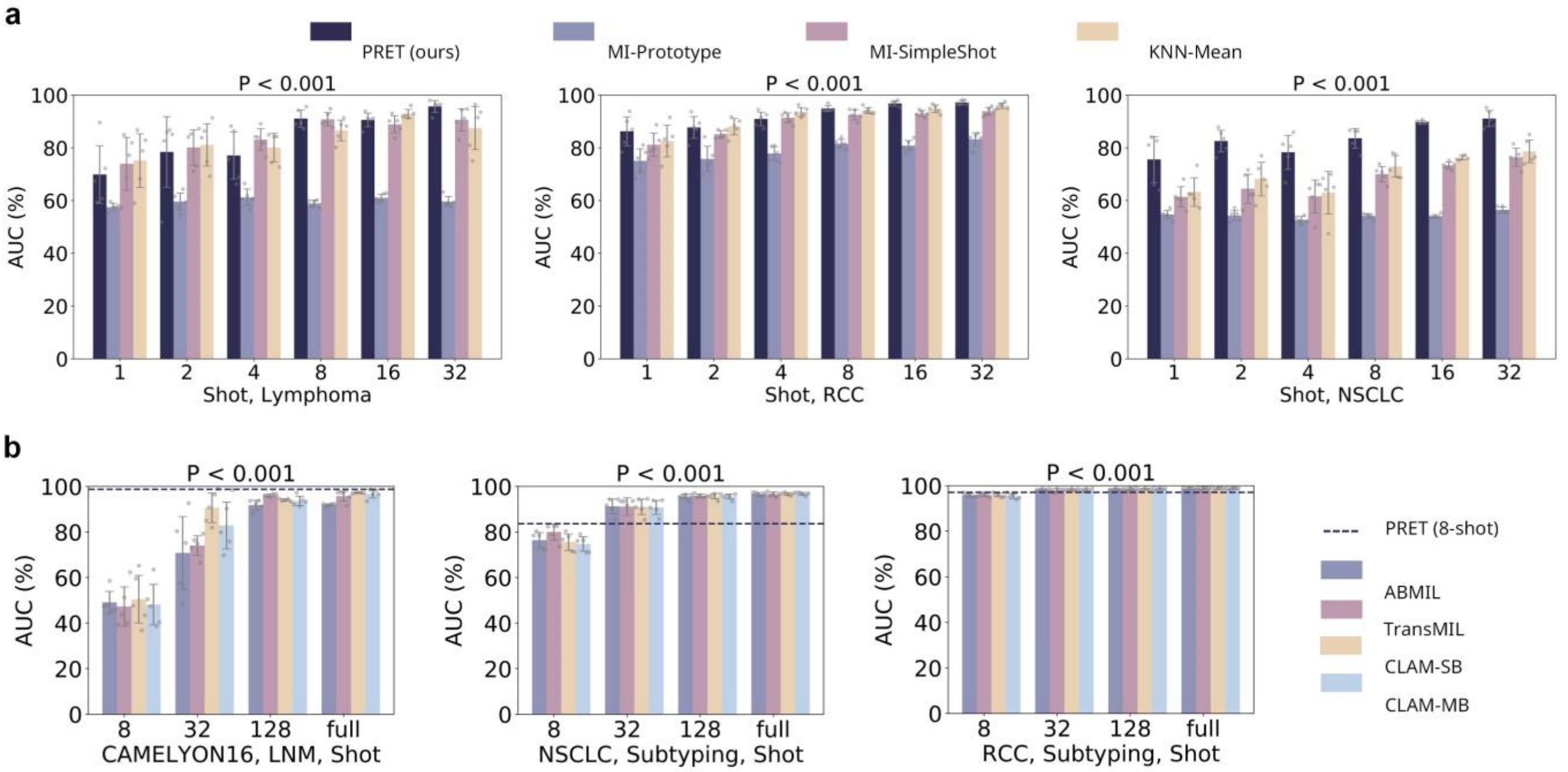

Extended Data Fig. 8 | PRET is scalable and achieves comparable performance to many-shot methods with much less data. a. PRET is scalable when applying more examples. We increased the example size (+ 16 and 32) and witnessed a continuous growth in performance. These results indicate that PRET is scalable and maintains its advantages over baseline methods ($p < 0.001$). The involved datasets are RCC, NSCLC, and Lymphoma under slide label to support more examples. b. PRET achieves comparable performance to many-shot fine-tuned methods using much less data (for example 8 vs. 128 or full). The many-shot methods are mainstream fine-tuned multiple instance learning methods, including ABMIL18, TransMIL37, CLAM-SB10, and CLAM-MB10. PRET applies only 8 examples per class (8-shot) without training, while many-shot methods are trained with much more data, including 128-shot and full data scale. The prompt type corresponds to Fig. 2b, and the calculation of p value follows the above principle. PRET archives higher results than many-shot methods at the same data scale for all three benchmarks. Moreover, our performance is higher than many-shot methods in the full data scale on the CAMELYON1628 lymph node metastasis detection task ($p < 0.001$), while achieving comparable results for the rest of the benchmarks ($p < 0.001$). a, b. The gray error bars indicate standard deviation across n = 5 independent experiments, and the gray dots are the specific results. The p value indicates the significance of PRET outperforming the best baseline with two-sided Wilcoxon test, reporting the median value across all repeat experiments and varied shots.